\RequirePackage{fix-cm}
\documentclass[twocolumn]{svjour3}          
\smartqed  

\usepackage{graphicx,calc}
\usepackage{hyperref}
\hypersetup{
    colorlinks=true,
    breaklinks=true,
    linkcolor=red,
    filecolor=magenta,      
    urlcolor=blue,
    citecolor=blue
}
\usepackage{pifont}
\usepackage{multirow}
\usepackage{subcaption}
\usepackage{amsmath}
\usepackage{bbm}
\usepackage{rotating}
\usepackage[dvipsnames]{xcolor}
\usepackage{comment}
\usepackage{booktabs}
\usepackage{float}
\usepackage[disable,colorinlistoftodos]{todonotes}

\usepackage{xcolor}
\definecolor{blue}{HTML}{4472C4}
\definecolor{green}{HTML}{70AD47}
\definecolor{orange}{HTML}{ED7D31}
\definecolor{purple}{HTML}{7030A0}
\definecolor{yellow}{HTML}{FFC000}
\definecolor{teal}{HTML}{00CC99}
\definecolor{pink}{HTML}{CC0099}

\newcommand{\cmark}{\ding{51}}
\newcommand{\cbox}[1]{\colorbox{#1}{\phantom{0}}}

\begin{document}
\title{Large Scale Genealogical Information Extraction From Handwritten Quebec Parish Records}
\titlerunning{Genealogical Handwritten Information Extraction}
%
\author{
Solène Tarride \and 
Martin Maarand         \and
        Mélodie Boillet  \and
        James McGrath \and
        Eugénie Capel \and 
        Hélène Vézina  \and
        Christopher Kermorvant
}


\institute{
            Solène Tarride  and Martin Maarand \at
              TEKLIA, Paris, France\\
              \email{starride@teklia.com}
             \and
            Mélodie Boillet and Christopher Kermorvant\at
              LITIS, Normandie University, Rouen, France \\
              TEKLIA, Paris, France \\
              \email{boillet@teklia.com kermorvant@teklia.com} 
            \and
            James McGrath, Eugénie Capel and Hélène Vézina \at
              BALSAC project, Université du Québec à Chicoutimi, Canada 
}

\maketitle              

\begin{abstract}

This paper presents a complete workflow designed for extracting information from Quebec handwritten parish registers. 
The acts in these documents contain individual and family information highly valuable for genetic, demographic and social studies of the Quebec population.
From an image of parish records, our workflow is able to identify the acts and extract personal information. The workflow is divided into successive steps: page classification, text line detection, handwritten text recognition, named entity recognition and act detection and classification. For all these steps, different machine learning models are compared. 
Once the information is extracted, validation rules designed by experts are then applied to standardize the extracted information and ensure its consistency with the type of act (birth, marriage, and death). This validation step is able to reject records that are considered invalid or merged.
The full workflow has been used to process over two million pages of Quebec parish registers from the 19-20th centuries. On a sample comprising 65\% of registers, 3.2 million acts were recognized. Verification of the birth and death acts from this sample shows that 74\% of them are considered complete and valid. These records will be integrated into the BALSAC database and linked together to recreate family and genealogical relations at large scale.




\end{abstract}

\section{The BALSAC project}

\subsection{From BALSAC to i-BALSAC}

For the last 50 years, the BALSAC project\footnote{\url{https://balsac.uqac.ca/}} has been building and consolidating a major database on the Quebec population. The core of the database is made of demographic events extracted from transcribed parish and civil registers. Birth, marriage and death records are linked together to reconstruct the Quebec population from the beginning of French settlement in the 17th century to the contemporary period. From the 1980’s, mostly marriage records were entered in the database, in order to concentrate on genealogical reconstructions used in genetic research. 

About ten years ago, the decision was made to add birth and death records and link them to the marriages already in BALSAC for a comprehensive coverage of families. During this time, it also became increasingly evident that the development of the database, which entails work on millions of records, could no longer rely exclusively on manual or semi-automatic operations. Fortunately, progress in machine learning opens up promising avenues for historical databases as handwritten text recognition (HTR) algorithms have improved significantly in the past few years. These considerations have led the BALSAC team to make the decision to rely on this technology for the transcription of Quebec civil registers and their integration in the database. 

These developments are part of an initiative supported by the Canadian Foundation for Innovation aimed at creating i-BALSAC, an infrastructure for the study of the Quebec population through a joint genealogical, genomic and geographic approach. This project, which will be completed in 2023, has three main components: the integration of demographic, genetic and geographical data, the development of statistical and mapping tools to enhance exploitation of the data, and the setting up of a web portal for access.

\subsection{Quebec civil and parish registers}

\begin{figure*}[t]
     \centering
     \begin{subfigure}[b]{0.49\textwidth}
         \centering
         \includegraphics[width=\textwidth]{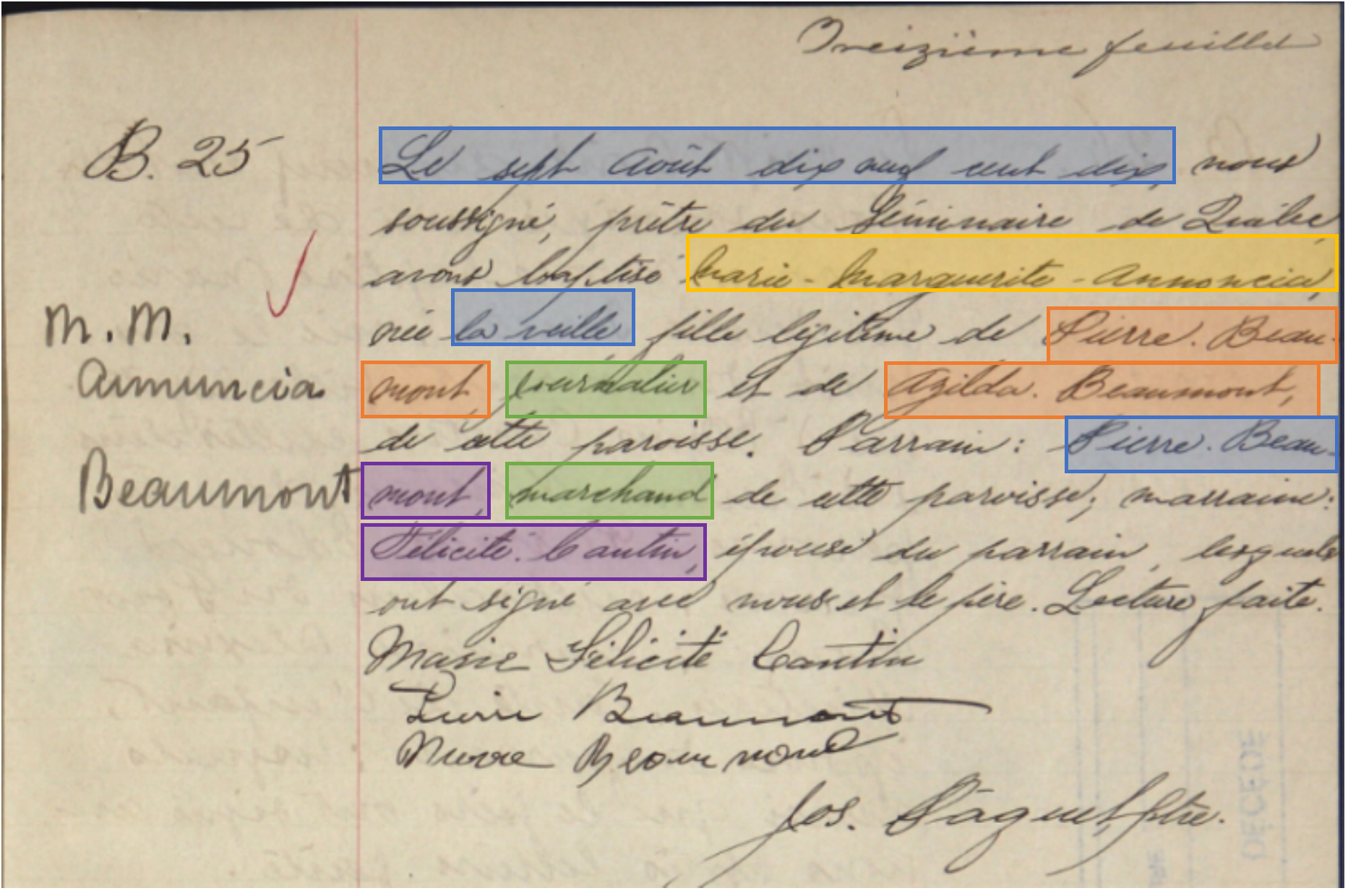}
         \caption{Act of birth}
         \label{fig:baptism}
     \end{subfigure}
     \hfill
     \begin{subfigure}[b]{0.49\textwidth}
         \centering
         \includegraphics[width=\textwidth]{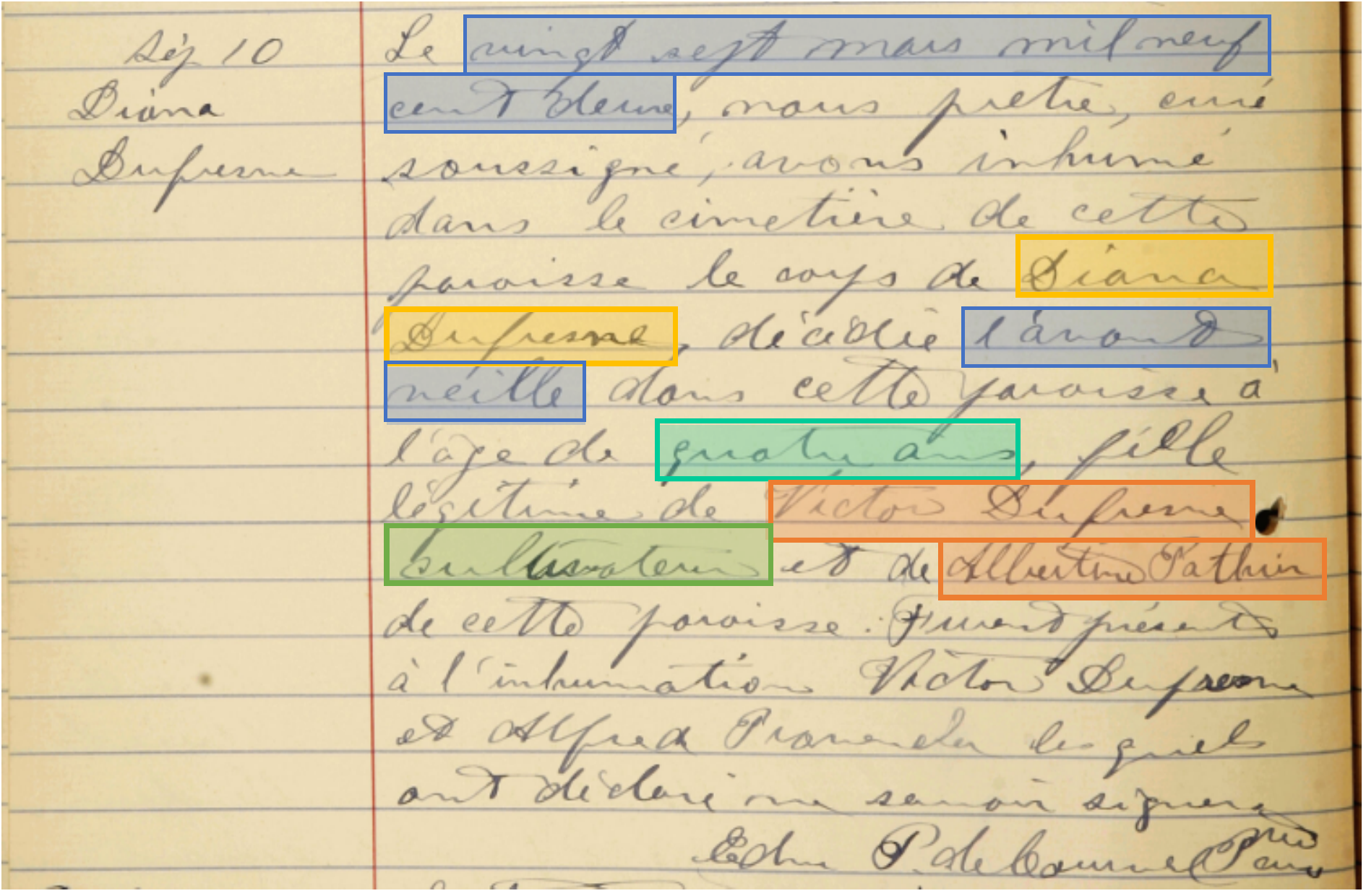}
         \caption{Act of death of a single person}
         \label{fig:death_married}
     \end{subfigure}
     \hfill
     \begin{subfigure}[b]{0.49\textwidth}
         \centering
         \includegraphics[width=\textwidth]{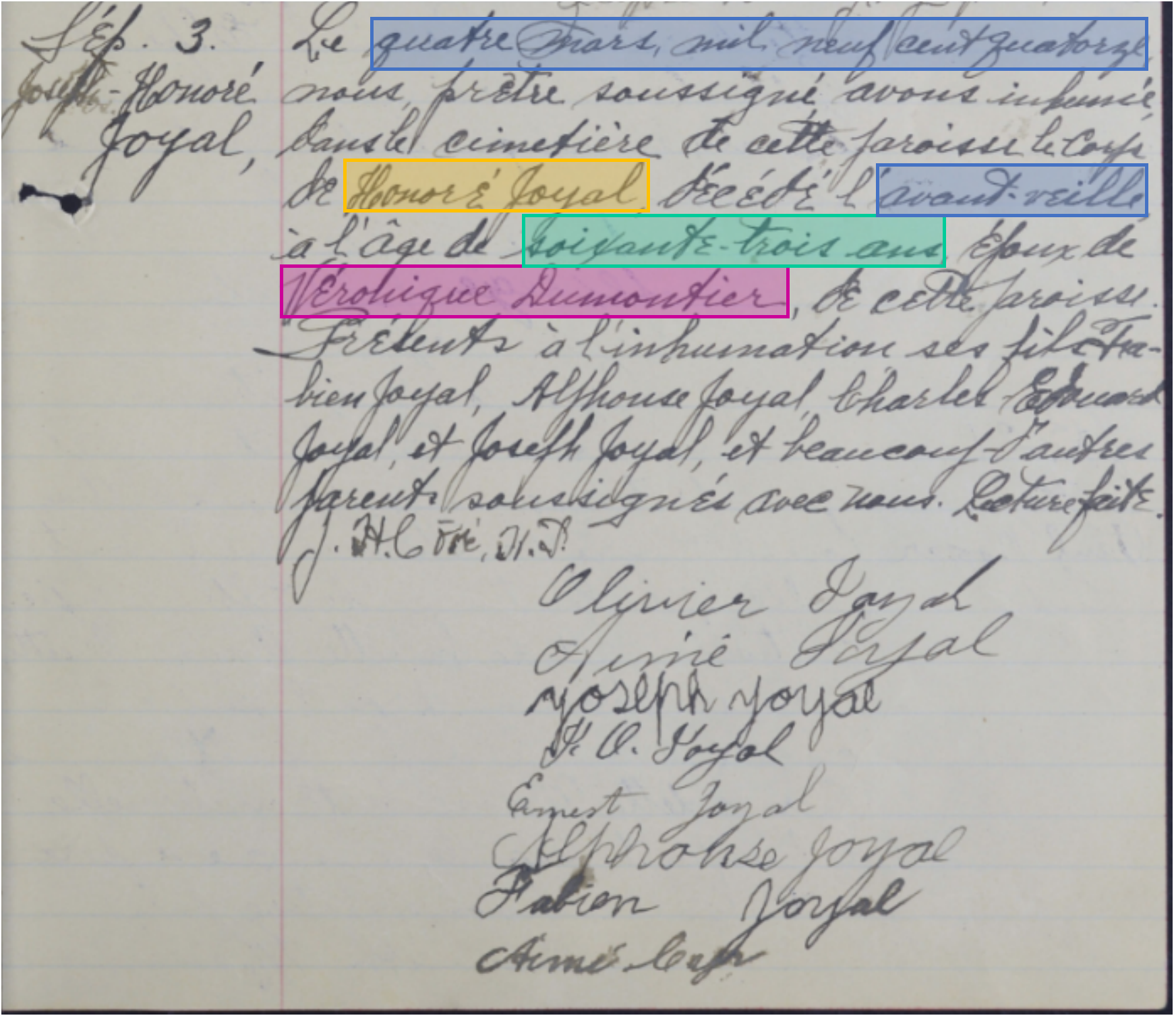}
         \caption{Act of death of a married person}
         \label{fig:death_single}
     \end{subfigure}
     \hfill
        \caption{Illustration of three acts in which important information is highlighted. \textbf{Legend}: \cbox{blue} Date;  \cbox{yellow} Subject of the act; \cbox{orange} Parents; \cbox{pink} Spouse; \cbox{teal} Age; \cbox{green} Occupation; \cbox{purple} Godfather/godmother.}
        \label{fig:register_act}
\end{figure*}

In the course of the i-BALSAC project, the birth and death certificates for the Quebec population between 1850 and 1916 will be integrated into the BALSAC database. High resolution images of these records were obtained thanks to a partnership with the Bibliothèque et Archives nationales du Québec (BAnQ). They amount to 1,995,579 digitized pages from 44,742 registers located in 1,985 different parishes.  Ultimately, the goal is to process these images to identify and index the various entities contained in each record such as names and surnames, dates, places, and occupations, as illustrated in Figure \ref{fig:register_act}. Since the integration of data includes a linkage procedure to connect information from various events pertaining to the same individual or the same family, the quality of the information, particularly on names and surnames, is of the utmost importance.

It is estimated that the complete processing of these images will make it possible to add approximately six million new records to the database, more than doubling its current size. Such an amount of data, covering the whole Quebec territory over seven decades, also means great variations in terms of page structure, handwriting styles, and multiple particularities across registers. Disparities in the data also comes from religious affiliation: about 60\% of the registers of the 1850-1916 period come from Catholic parishes, while the remaining 40\% are composed of non-Catholic confessions, including Protestants and their various denominations as well as Jews and Orthodox. These figures are slightly misleading however since registers are not organized the same way across religions: comparing Catholics and non-Catholics in terms of number of images, the proportions change to 78\% and 22\% respectively. In addition, most Catholic registers are in French while most non-Catholics are written in English
Such a diversity is a challenge for deep learning techniques that are known to require training data representative of the full target corpus in order to be effective. 
The selection and annotation of training data is described in Section \ref{sec:annotated-ground-truth}.

In this article, we introduce three main contributions:
\begin{itemize}
    \item A complete workflow for nominative information extraction in historical registers, from the image to the database integration ;
    \item A comparison of multiple machine learning models at each step of the workflow, allowing us to identify the main challenges for information extraction ;
    \item Rules designed for content verification and standardization, and an unsupervised metric for quality estimation.
\end{itemize}

The paper is organized as follows. 
First, an overview of the main approaches for information extraction from digitized documents is presented in Section \ref{sec:biblio}. 
In section \ref{sec:workflow}, we describe the training data and present each step of the workflow proposed for information extraction. For each step, we compare multiple machine learning models and discuss the results. 
In section \ref{sec:validation}, we describe the procedure for content validation and standardization designed to ensure that the right information is entered into the BALSAC database.
Finally, we discuss the results and propose some insights in Section \ref{sec:conclusion}.

\section{Related works}
\label{sec:biblio}
\todo[inline]{Uniformiser les citations (ajouter Machin et al.)}
Recent improvements in computer vision and natural language processing have led to significant advances in the field of automatic document understanding. 
At the same time, many systems have been developed to ease ground truth production \cite{transkribus,diva2016}, allowing researchers to train complex and efficient models for information extraction from full-page documents. 

\subsection{Information extraction from full pages}

There are two main approaches to perform information extraction from full pages:
\begin{itemize}
    \item \textbf{Single-stage workflow}: information extraction is tackled in a single stage by combining features extracted from various modalities: image, text, layout. 
    \item \textbf{Step-by-step workflow}: information extraction is divided into multiple successive steps: page classification, text detection, text recognition, and named entity recognition. 
\end{itemize}

\subsubsection{Single-stage workflow} 
Several single-stage models have been proposed to extract relevant information in a single stage by taking advantage of both visual and textual features. 

Single-stage methods usually rely on pre-trained models to extract image, text and layout embeddings from full pages. These embeddings are then combined and used in the decoder to predict semantic labels \cite{layoutlm,douzon2022}. 
Some methods also take advantage of a graph learning module that models the relationships between nodes and generates a richer, more structured representation of the extracted information \cite{Yu2021PICKPK,gcnn_ie}.
Other methods extract generic features that are shared by multiple branches dedicated to specific tasks, such as text zone detection, text zone recognition and information extraction \cite{Carbonell2020,Wang2021}. 

Single-stage methods are very efficient as they combine multiple document modalities (image, text, layout) for information extraction. Such approaches are especially well-adapted to structured documents, such as tickets, invoice or receipts, in which semantic information can be derived not only from the textual content, but also from the layout. For example, in a census document, names will always be located in a dedicated column, thus in a specific area of the page. Additionally, performing information extraction in one stage is interesting to avoid error accumulation (in a traditional workflow, bad text line detection will have an impact on the text recognition). 
However, they are not particularly suited to weakly structured documents, in which semantic information is mostly derived from the text. Additionally, single-stage workflows rely on pre-trained models that are often trained on printed documents or synthetic documents, and therefore not directly usable for historical or handwritten documents.

\subsubsection{Step-by-step workflow}
Another way of extracting relevant information from full pages consists in dividing the workflow in successive, more simple steps: page classification, text detection, text recognition, named entity recognition, act detection and classification. The final workflow can be built using successive state-of-the-art components for each task. Each individual task has been extensively studied in the literature, so in this study, we mainly focus on models applied to historical documents. 

\paragraph{Page classification.} Page classification of historical documents usually consists in associating each page with a class that describes the period of the document, its place of origin, the script used or the author of the document \cite{das-pageclassification-comp-2022,das-pageclasification}. In our case, the classification task would help to discard pages without any act (cover, blank page, index…). As a result, it could be tackled as an outlier detection problem, as the dataset is extremely unbalanced, and few pages without act have been annotated.

\paragraph{Text line detection.} Many open-source models have been proposed to tackle text line detection from historical documents, mainly ARU-Net \cite{gruning2018}, dhSegment \cite{dhsegment}, DocExtractor \cite{monnier2020docExtractor} and Doc-UFCN \cite{boillet2021}. These models have been pre-trained on multiple datasets \cite{diva2016,readbad2017}, making them very efficient on a wide variety of historical documents. They can be used easily out-of-the-box, or fine-tuned on a specific set of documents. 

\paragraph{Handwritten text recognition.} Many open-source architectures have been proposed for handwritten text recognition (HTR). This task is generally performed at line-level, either relying on a CNN-HMM architecture \cite{kaldi}, a CNN-RNN architecture \cite{PyLaia}, or a transformer-based architecture \cite{dan}. Recently, an increasing number of articles have been dedicated to learning from paragraphs or full pages \cite{bluche2016,wigington2018,dan}.

\paragraph{Named entity recognition.} Open source NER models mainly rely on Transformer architecture: CamemBERT \cite{camembert}, spaCy \cite{spacy}, Stanza \cite{qi2020stanza} or FLAIR \cite{akbik2018coling}. Two recent surveys compare open-source NER libraries \cite{monroc2022-ner-survey,abadie2022-ner-survey} on noisy text automatically recognized from historical documents. Both studies conclude that most NER systems achieve high-performance results, even when trained on few training data. They also highlight that errors accumulate a lot in such as sequential workflow, as text detection and text recognition have a significant impact on named entity recognition results.

\paragraph{Joint text and named entity recognition.} Alternatively, handwritten text recognition and named entity recognition can be performed jointly either at line-level \cite{carbonell2018,tarride2022}, or at paragraph-level \cite{InstaDeep-JointHTRNER}. These models predict a special semantic token before each word of interest, such as \textit{$<$name$>$}, \textit{$<$date$>$}, or \textit{$<$occupation$>$}. Most of these approaches have been proposed as part of the 2017 Information Extraction in Historical Handwritten Records (IEHHR) competition\footnote{\url{https://rrc.cvc.uab.es/?ch=10&com=introduction}} \cite{IEHHR2017}. Following the same principle, attention-based neural networks such as DAN \cite{dan} could also be used for information extraction from full pages, by taking advantage of special tokens for both semantic labelling and structure description.

\paragraph{Act detection.} Act detection is often performed using only visual features. For example, signatures and first text lines are automatically detected using a CNN and combined to localize the acts \cite{tarride2021}. In \cite{capobianco2019}, records are counted using a CNN trained on synthetic documents that are generated with a similar layout.
In cases where the visual features are not sufficient to correctly detect the acts, some works have focused on combining visual and textual features. This has been achieved by using Probabilistic Index Map \cite{prieto2020} or by enriching the image with the localization of first text lines that are detected using their textual content \cite{boillet2021}.

\subsection{Large-scale workflows for automatic document understanding}

The BALSAC project aims to extract information from millions of handwritten historical records. 
Beyond the scientific challenge of automatic information extraction, processing millions of pages is also a technical challenge with multiple dimensions: data management, computing resources, and processing time.

Few works have been dedicated to information extraction in historical documents at such a large scale. Some of them also implemented rules for unsupervised content verification.
A first approach was introduced in 2013 to perform information extraction in real conditions and at a large scale on eight million pages from the 1930 US census \cite{nion2013}. For this task, the authors report a 70\% automation rate: 30\% pages were manually processed as the automatic system was considered unreliable.
Other researchers focused on template-based information extraction on printed books \cite{embley2018}. The text was automatically recognized using an OCR, and semantic labelling was performed using template matching. The authors processed several hundred thousand pages using this approach.
Another approach was introduced for lexicon-free keyword spotting on hundreds of thousands of German historical registers \cite{lang2018}. The authors managed to support structured queries for information extraction in complex tabular documents.
The SYNTHESIS+ project \cite{walton2022} focused on developing a cloud-based platform to extract information from images of natural history specimens at large scale. To do this, the authors designed a workflow that integrates various open source tools, mainly for object detection, image segmentation, and text recognition.
More recently, information extraction has been performed from hundreds of thousands pages of the early 20th century Paris Census \cite{constum2022}. The authors took advantage of the tabular structure to implement rules for normalization and content verification.


\subsection{Discussion}

The i-BALSAC project began in 2019. At this time, single-stage methods for information extraction were still in their early stages. Since then, machine learning models have evolved a lot, and single-stage methods have proven to be very efficient. If the project were to start again now, we would surely consider developing a single-stage workflow. 

That being said, we decided to tackle information extraction with a step-by-step workflow. This approach is very convenient to divide the workload among the team. Moreover, it allows us to identify and focus on the most difficult steps. Finally, dividing the workflow into different modules is practical to maintain, as each module can be updated independently. 

\section{Genealogical information extraction workflow}
\label{sec:workflow}
\begin{figure*}[t]
    \centering
    \includegraphics[width=\textwidth]{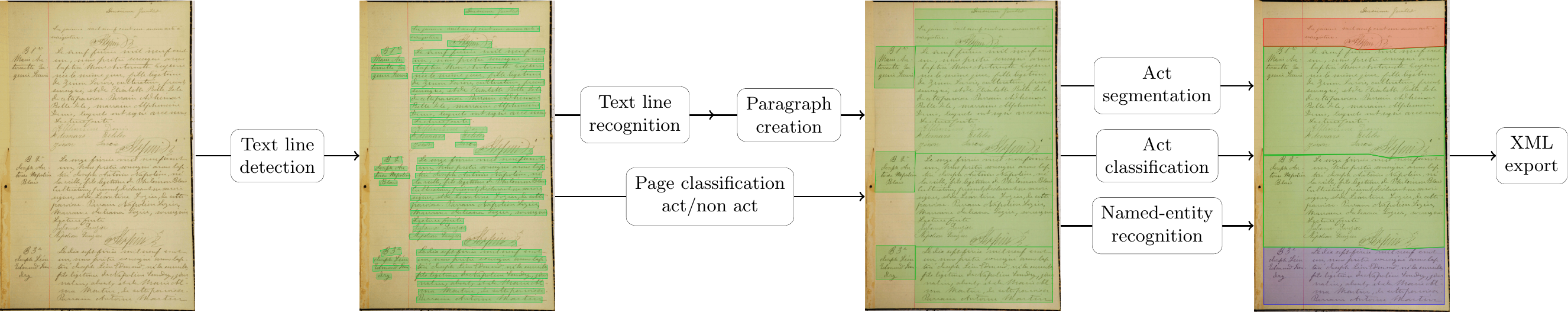}
    \caption{Overview of the workflow. First, text lines are detected. Then, the text is automatically transcribed and paragraph are created based on visual features. At the same time, each page is classified as containing act information or not. Next, acts are localized and classified based on their position in the page (in red if they start on the previous page, in green if they start and end in the current page, in purple if they end on the next page), and automated named entity recognition is performed. Finally, complete recognition results are exported in an XML file.}
    \label{fig:overview}
\end{figure*}

The complete workflow for the extraction of personal information from scanned registers is presented in Figure \ref{fig:overview} and described in this section. The first step consists in detecting the text lines. Then, for each page, text lines are used to classify the type of page (act or non act) and for handwritten text recognition (HTR). Recognized text lines are then grouped into paragraphs and the corresponding text is used for name-entity recognition (NER). The text of the lines is also used to detect the boundaries of the acts as well as their types (birth, marriage, death). Finally, all the extracted information is exported into an XML file.

As it is well known in document processing that the errors accumulate along the different steps, we design the workflow in order not to take early hard decisions. For example, we process all the pages with the complete workflow, even if they are predicted as not containing any act information. In the same spirit, we wait for the end of the complete recognition workflow to reconstruct the hierarchical structure between text lines, acts, and paragraphs. The grouping is performed based on text line localization and content, in order to make the decision more reliable.

Most of the steps of the processing are based on machine learning algorithms and thus need annotated data to be trained. A description of the training data is given in the following section.

\subsection{Creation of the ground-truth data}
\label{sec:annotated-ground-truth}

To address the challenge of data variability, the decision was made to produce training data independently of religious affiliation and language. 
The ground truth for training the different models of the extraction workflow consists of 913 transcribed and annotated images picked from 74 manuscript registers. 
They have been selected in order for the sample to be representative, based on the BALSAC team's knowledge of the records.
The statistics of the annotated dataset at different levels (page, act, lines, words) are given in Table \ref{tab:dataset_split}. 

\begin{table*}[t]
\centering
\caption{Description of the dataset used in our experiments: counts of images, acts, lines, and words are provided for each subset. The distribution of the acts classes is also detailed}
\label{tab:dataset_split}
\begin{tabular}{lrrrrrrrrr}
\toprule
& \multirow{2}{*}{\textbf{Images}} & \multirow{2}{*}{\textbf{Pages}} & \multicolumn{5}{c}{ \textbf{Acts}} & \multirow{2}{*}{\textbf{Lines}} & \multirow{2}{*}{\textbf{Words}} \\
&  &  & \textit{\textbf{start}} & \textit{\textbf{center}} & \textit{\textbf{end}} & \textit{\textbf{full}} & \textit{\textbf{total}} &  &  \\
\midrule
\textbf{train} & 730 & 864 & 504 & 2 & 488 & 1,476 & 2,470 & 36,941 & 167,166 \\
\textbf{val} & 92 & 107 & 66 & 1 & 58 & 181 & 306 & 4,592 & 20,947 \\
\textbf{test} & 91 & 106 & 62 & 0 & 52 & 176 & 290 & 4,323 & 19,884 \\ 
\midrule
\textbf{Total} & 913 & 1,077 &632 & 3 & 598 & 1,833 & 3,066 & 45,856 & 207,997 \\
\bottomrule
\end{tabular}
\end{table*}

The transcription and annotation were done using Transkribus\footnote{\url{https://readcoop.eu/transkribus/}} in three steps: layout analysis, transcription, and entity tagging.

The detection of text lines and text regions was performed using Transkribus's CITLab tool, with manual corrections and additions. The “marginalia” zone was annotated with a special label. The acts were also manually annotated, as they are the key unit in these registers. Bounding boxes of the acts were drawn so that they cover the main text of the act as well as the marginalia. 
Since an act can be written over more than one page, we also tagged each act as either:
\begin{itemize}
    \item \emph{act start}, to indicate that the act ends on the following page ;
    \item \emph{act center}, to indicate that the act starts on the previous page and ends on the following page ;
    \item \emph{act end}, to indicate that the act starts on the previous page ;
    \item \emph{full act}, to indicate that the act starts and ends on the current page.
\end{itemize}

Table \ref{tab:dataset_split} also describes the distributions of these elements in the annotated dataset.

The text lines were manually transcribed by research assistants, who have also tagged the main named entities, as detailed in Table \ref{tab:entity_stats}. 
The named entity recognition (NER) task was limited to dates, names, occupations and declared residences because these entities are found consistently in these registers. 
Moreover, these entities are easier to retrieve than other pieces of information, as they span over only a few words. 
In contrast, complex entities such as attendance (whether a person mentioned in the record was physically present for the act's recording) or signature information are often part of complex sentences, and can require interpretation. 

Automated named entity recognition is crucial for this project, as the entities help to locate events in time and in space, specify the role of individuals mentioned in the records, and provide socioeconomic information such as occupations, honorific titles, and literacy. Furthermore, names, places, and occupations can be linked to existing dictionaries containing all the variations previously identified in the BALSAC database over more than three centuries of records.
Lastly, but most importantly, the selected entities are those required for the family and population reconstruction. Event records are linked based on names, but context variables such as places and even occupations can help find the right individuals or families.

Entity tagging thus serves a dual objective: it helps validate the transcription through an entity-oriented proofreading operation, and as ground truth for named entity recognition it will help to provide an overview of the textual structure of the registers. Moreover, standardized tags will help to answer questions about the consistency of the textual structure of registers over time, to document the structure of records over a long period, and to compare ecclesiastical and civil registration requirements to demographic data from other sources.

\begin{table}[t]
\caption{Distribution of the named entities present in the annotated dataset}
\label{tab:entity_stats}
    \centering
    \begin{tabular}{lrrrrr}
    \toprule 
    & & \textit{\textbf{PER}} & \textit{\textbf{LOC}} & \textit{\textbf{DATE}} & \textit{\textbf{OCC}}\\ 
    \midrule
    \textbf{Count} & & 15,810 & 2,823 & 4,551 & 2,380\\
    \multirow{2}{*}{\textbf{Avg length}} & \textit{words} & 2.33 & 2.08 & 5.21 & 1.31\\
    & \textit{chars} & 14.27 & 12.33 & 23.28 & 10.44\\ 
    \bottomrule
    \end{tabular}
\end{table}

\subsection{Workflow management} 

Managing a large-scale digital document processing project requires the implementation of a workflow management system to execute each step and ensure communication between processes. The system must also be able to distribute tasks according to the available computing capacity. The system must be based on a technology that is reliable, sustainable and cost-effective. 

For the management of the 2 million images, representing 7.2 To of data, we have chosen a solution implementing the IIIF protocol. The images, initially in TIFF, were converted into JPEG both to reduce their size and to be compatible with IIIF image servers. All the images are made available using a Cantaloupe\footnote{\url{https://cantaloupe-project.github.io/}} IIIF server, allowing to access both the full images and portion of images according to the IIIF image API\footnote{\url{https://iiif.io/api/image/2.1/}}.

For the management of the documents and of the processing results, we have based the whole project on the Arkindex platform\footnote{\url{https://doc.arkindex.org/}} both for storing the processing results and distributing the tasks on a cluster of processing servers. Arkindex also includes a visualization interface to analyse the cause of failures.

\todo[inline]{@CKE Citer Walton et al. \cite{walton2022} dans une phrase ou commenter la ligne}

\subsection{Page classification (containing/not containing acts)}

\begin{figure*}[t]
    \centering
    \includegraphics[width=\textwidth]{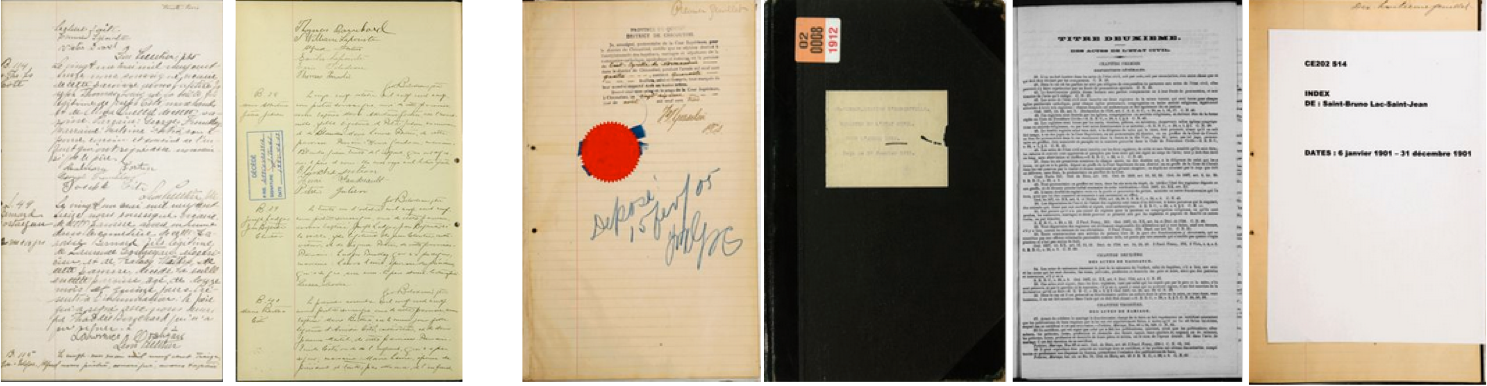}
    \caption{Page types in the BALSAC corpus: two pages containing acts on the left, four pages containing no act on the right. Pages with no acts are much more heterogeneous than pages containing acts.} 
    \label{fig:act_non_act}
\end{figure*}

Over the 2 million pages, not all of them contain acts: some are blanks pages or covers, some are printed instructions from the registers, and some are indexes. 
As presented on Figure \ref{fig:act_non_act}, pages with acts are very homogeneous, whereas other pages are very heterogeneous. 

Page classification is an important step for quality control. For example, if no act is found on a page that was classified as containing act information, then there might be an issue with handwritten text recognition, named entity recognition or act detection. 
Since only act pages were available in the annotated corpus, it was decided to detect non act pages as outliers compared to act pages. The page classification algorithm is described and evaluated in the following paragraphs. 







\subsubsection{Outlier detection models and features}

After preliminary experiments, we selected two algorithms for outlier detection from Scikit-Learn's library \cite{scikit-learn}: Isolation Forest \cite{liu2008isolation} and Local Outlier Factor \cite{breunig2000lof}.

\paragraph{Isolation forest} combines several isolation trees to identify abnormal data points. The idea of this algorithm is that outliers are easier to isolate as they are closer to the root node, leading to a smaller average path length as compared to inliers. As a result, the path length is a measure of normality, and can be used to calculate an anomaly score. Data points with an anomaly score higher than 0.5 are considered as outliers. 

\paragraph{Local Outlier Factor} is based on the estimation of the local density of neighbors for each point. The local density of each point is compared with the local densities of its neighbors: if the neighbors' densities are much higher, then the point comes from a sparser region and is considered as an outlier.\\

\begin{figure*}[t]
    \centering 
     \includegraphics[width=0.235\textwidth]{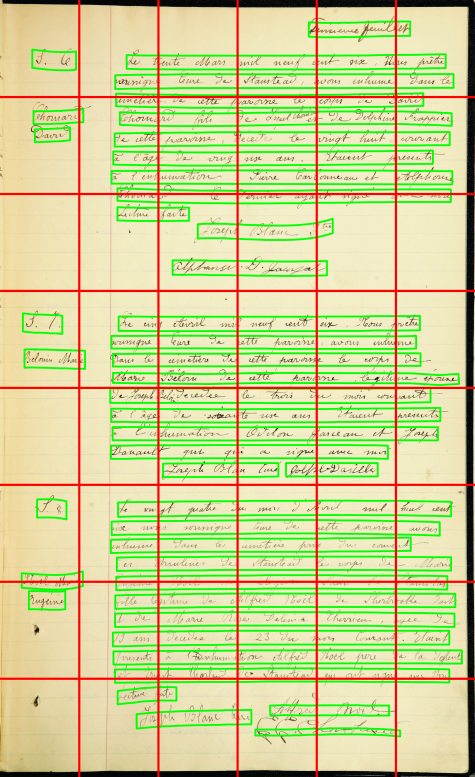}
    \includegraphics[width=0.1\textwidth]{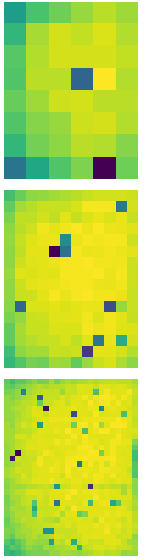}
    \includegraphics[width=0.098\textwidth]{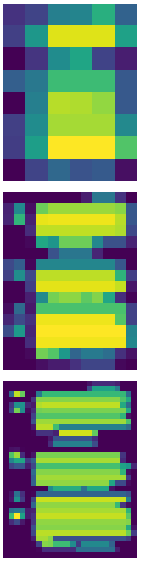}
    \includegraphics[width=0.1\textwidth]{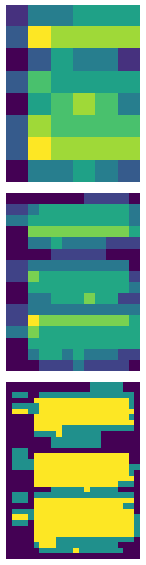}
    \caption{Comparison of different features for outlier detection. From the input image (left), three different features are presented with three different grid sizes (right). From left to right: subsample image, line density grid, line count grid. From top to bottom: different grid sizes: (8$\times$6), (16$\times$12), (32$\times$24)} 
    \label{fig:extracted_feature_comparison}
\end{figure*}

For these two algorithms, we compared three different types of features:
\begin{itemize}
    \item Subsampled image projections: the input image is subsampled to a size $rows \times columns$, then the column and row projections are computed and transformed in a vector of size $rows + columns$.
    \item Line density grid: a grid of $rows \times columns$ is defined on the input image. For each cell, the percentage of text lines area is computed, transformed into a vector and used as features.
    \item Line count grid: a grid of $rows \times columns$ is defined on the input image. For each cell, the number of lines intersecting the cell is computed, transformed into a vector and used as features. 
\end{itemize}
An illustration of the different features extracted from a page is presented on Figure \ref{fig:extracted_feature_comparison}.

\subsubsection{Outlier detection results}

Two main layouts can be found in the corpus, depending on the scanning method: portrait (single page) and landscape (double page). We found that a model trained on both portrait and landscape orientations was unable to work properly, since the features learned on portrait images were "stretched" for landscape images. 
Therefore, a model was trained only on single images, as landscape pages can be easily split into two single pages. We used 739 single page images from the annotated dataset to train the outlier detector. 

To evaluate the models trained in an unsupervised manner, 20 images containing acts and 20 images without any act were annotated from unseen data. Grid search over different feature matrix/vector sizes and some hyperparameters for Isolation Forest and Local Outlier Factor was used to find the best model. In the end, Isolation Forest that used the line count grid of size (8$\times$6) got the best F1-score and was selected.


The best model was tested on registers from the region of Saguenay. 100 pages were sampled in a stratified way (78 containing acts, 22 without any act) to evaluate the error rate on unseen data.
The evaluation results are presented in Table \ref{tab:outlier_classifier_evaluation}. 
We obtained a combined F1-score of 97\% which was satisfying and most importantly a recall of 100\% on images containing acts.
Additionally, we applied this model to all the images from the region of Saguenay to validate the results visually. Prediction results are presented in Table \ref{tab:outlier_classifier_prediction}, and seem consistent with the evaluation results.

\begin{table}[t]
\centering
\caption{Evaluation results of the outlier detector on 100 annotated images extracted from the registers of the region of Saguenay}
\label{tab:outlier_classifier_evaluation}
\begin{tabular}{lrrrr}\toprule
& \textbf{Precision} & \textbf{Recall} & \textbf{F1} & \textbf{Support} \\
\midrule
\textbf{Act} & 0.96 & 1.00 & 0.98 & 75 \\
\textbf{No act} & 1.00 & 0.88 & 0.94 & 25 \\
\midrule
\textbf{Total/Average} & 0.97 & 0.97 & 0.97 & 100 \\ \bottomrule
\end{tabular}
\end{table}

\begin{table}[t]
\centering
\caption{Prediction results of the outlier detector on 36,700 images extracted from the registers of the city of Saguenay}
\label{tab:outlier_classifier_prediction}
\begin{tabular}{lrr}\toprule
& \textbf{Count} & \textbf{Percentage} \\
\midrule
\textbf{Act} & 28,478 & 77.6\% \\
\textbf{No act} & 8,222 & 22.4\% \\
\midrule
\textbf{Total} & 36,700 & 100.0\% \\ \bottomrule
\end{tabular}
\end{table}

\subsection{Layout analysis and text line detection}
\label{sec:layout-analysis}

\subsubsection{Text line detection model}

For the detection of text lines, we compared three state-of-the-art systems: Doc-UFCN \cite{boillet2020}, dhSegment \cite{dhsegment} and ARU-Net \cite{gruning2018}. They have been selected for their high performances on various historical document segmentation tasks. 

\textbf{Doc-UFCN} has been chosen since it has yielded good results on various historical datasets for the text line detection and for its ability to be trained on a reduced amount of training images. It aims at a general and flexible approach: depending on the input data, it can address, for example, text line extraction or complete layout analysis. Doc-UFCN is a Fully Convolutional Network (FCN) comprised of (1) a contracting path composed of dilated convolutions that allow to learn an input image representation while augmenting the receptive field, and (2) an expanding path that is trained to output pixel probability maps.

\textbf{dhSegment} has also shown high performances on multiple historical document segmentation tasks. dhSegment is also a Convolutional Neural Network (CNN) that follows an encoder-decoder architecture, where the encoder is a ResNet \cite{resnet} model pre-trained on natural scene images \cite{imagenet}. The decoding path is similar to Doc-UFCN's one.

\textbf{ARU-Net} has been chosen for its good performance mainly for the baseline detection task. ARU-Net is also the text line detection model integrated into Transkribus \cite{transkribus}, the most popular platform for historical document processing. It is an extended version of the standard U-Net that comprises spatial attention and a residual structure.

The three systems have been trained on a wide set of diverse historical public and private datasets, including annotated BALSAC data, with the train / validation / test sets exposed in Table \ref{tab:dataset_split}. For this historical line model, two classes were defined: the background and the text line. The training details and results are discussed in \cite{boillet2021}.

\subsubsection{Text line detection results}

The trained models are evaluated using the Intersection-over-Union (IoU) metric, which compares the areas of the expected ground truth zones and the predicted ones. In addition, the predictions are evaluated at object level using Average Precision (AP) values. Based on a IoU threshold, the AP allows knowing the amount of correctly predicted text lines. Table \ref{tab:textline_segmentation} presents the IoU results as well as AP values for two thresholds (50\% and 75\%) and the mean AP (thresholds ranging from 50\% to 95\%).


\begin{table}[t]
\centering
\caption{Text line detection results obtained by Doc-UFCN, dhSegment and ARU-Net models trained on various historical datasets and evaluated on BALSAC test images}
\label{tab:textline_segmentation}
\begin{tabular}{lrrrr}
\toprule
& \textbf{IoU} & \textbf{AP@.50} & \textbf{AP@.75} & \textbf{mAP} \\ \midrule
\textbf{Doc-UFCN} & 0.87 & \textbf{0.98} & \textbf{0.91} & \textbf{0.76} \\
\textbf{dhSegment} & 0.74 & 0.94 & 0.54 & 0.51 \\
\textbf{ARU-Net} & \textbf{0.98} & 0.76 & 0.20 & 0.34 \\
\bottomrule
\end{tabular}
\end{table}

The IoU values obtained by Doc-UFCN and ARU-Net, higher than 85\%, are better than what we usually get using diverse datasets for training. When looking at the AP values, we see that Doc-UFCN model clearly outperforms the two other systems. Indeed, the AP values obtained by Doc-UFCN show that the predicted objects are well predicted: a good localization and really low number of predicted merged lines. 

When dealing with a large corpus such as BALSAC, processing time is an issue that must be optimized. Since the development of the complete workflow is incremental, images are processed several times so that the complete workflow is run several millions of times. Large deep neural networks are known to be computationally expensive. As an example, we compared the processing time of Doc-UFCN, dhSegment and ARU-Net on a GPU GeForce RTX 2070 8G. On average, dhSegment took 2.95 seconds per page and ARU-Net 1.39 seconds, where Doc-UFCN only took 0.41 seconds. Being 7 times faster than dhSegment, Doc-UFCN allows processing 2 million images in 9.5 days instead of 66 days.

According to the results obtained by the three trained models, the Doc-UFCN model has been used to detect the lines on BALSAC images. We not only chose this model for its good performance for the line detection task, but also for its competitive inference speed compared to other state-of-the-art systems.

\subsubsection{Unsupervised line detection metric} 

The Doc-UFCN text line detection model produces a mean confidence score along with the predicted lines. However, it is known that modern deep neural networks are poorly calibrated \cite{guo2017} and that their confidence score cannot be trusted, particularly for the Doc-UFCN model whose predictions are at pixel level, whereas we need a confidence score at text line level. An external quality metric on the text line detection process had thus to be developed.

To make it easier to find images where the line detection is poor, we created a simple metric to compare the detected line heights with the median line height of that image. The ratio of line heights that are half bigger (or smaller) than the size of the median is used as a metric reflecting the line detection quality. In other words, if the detected line heights deviate a lot, then there is probably a problem with the detection. 

Let $H$ be the list of the predicted lines on an image, $\tilde{h}$ the median of the heights of lines in H and $l$ a single line from that list having a height $h$. If, for most of the lines, their heights are similar so that $\alpha \times \tilde{h} \leq h \leq (1+\alpha) \times \tilde{h}$, with $\alpha \in [0,1]$, then the line detection was probably correct; otherwise,  there might be an issue. Therefore, the line detection quality ratio is calculated in the following way:
\begin{equation}
  Q_{line} = \frac{1}{|H|} \sum_{h \in H} \mathbbm{1}_{ h \in [\alpha \times \tilde{h} , (1+\alpha) \times \tilde{h}]}
\end{equation}

This metric allows to correctly distinguish the correct from the bad text line detections. This is possible on BALSAC images since one page is usually written by one person (except for the signatures) and there are no headings with a bigger font size in the acts. More standard ways for finding bad lines, based on standard deviation, were tried too, but they were not good enough. 

Based on the ratios and using $\alpha=0.5$, we defined 5 classes 
($\leq1\%$, $1-5\%$, $5-25\%$, $25-50\%$, $>50\%$) 
where the name of the class describes the ratio of bad lines predicted on an image. These classes help us find images with bad line detections that will be used for manual correction and active learning.
The distribution of bad line classes on Saguenay test set is described in Table \ref{tab:bad_line_classes_distribution}.
We can see that 90\% of the documents have a bad line ratio lower than 5\% which seems satisfying.

\begin{figure*}[t]
    \centering 
    \includegraphics[width=0.182\textwidth]{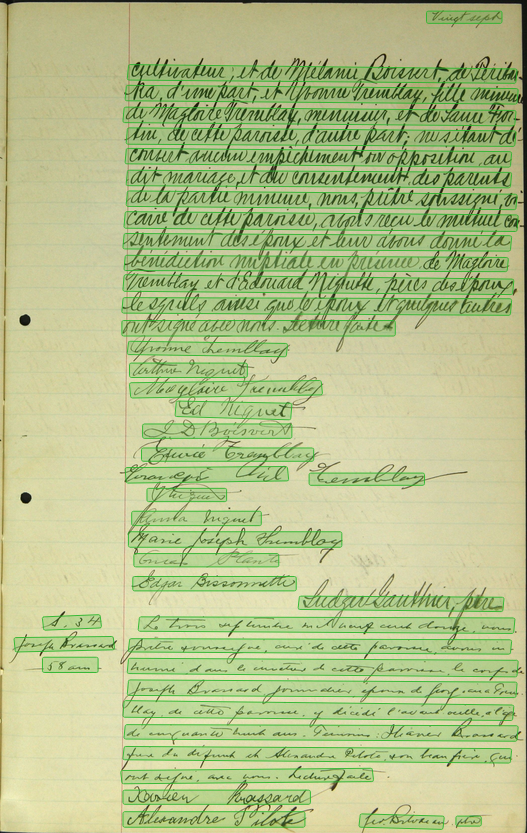}
    \hfill
    \includegraphics[width=0.18\textwidth]{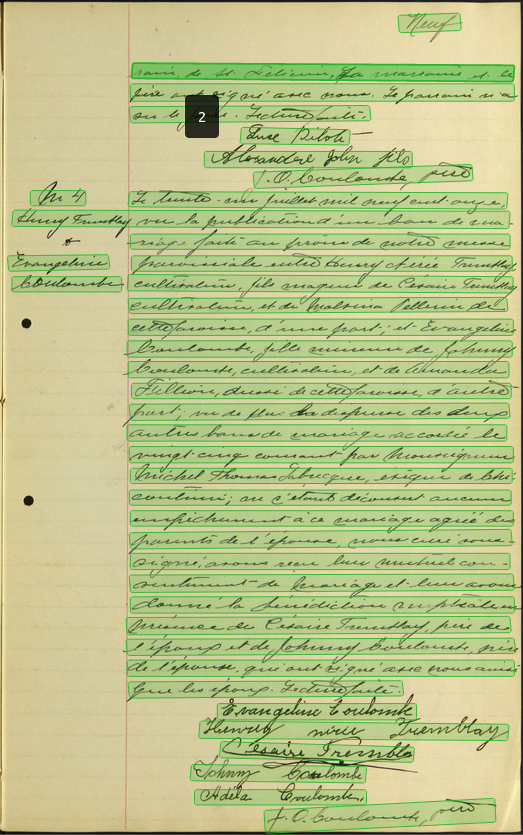}
    \hfill
    \includegraphics[width=0.19\textwidth]{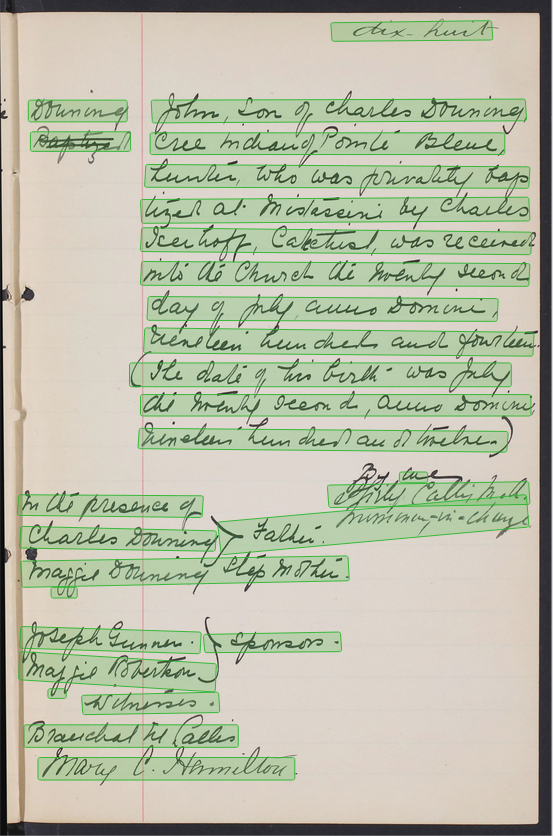}
    \hfill
    \includegraphics[width=0.175\textwidth]{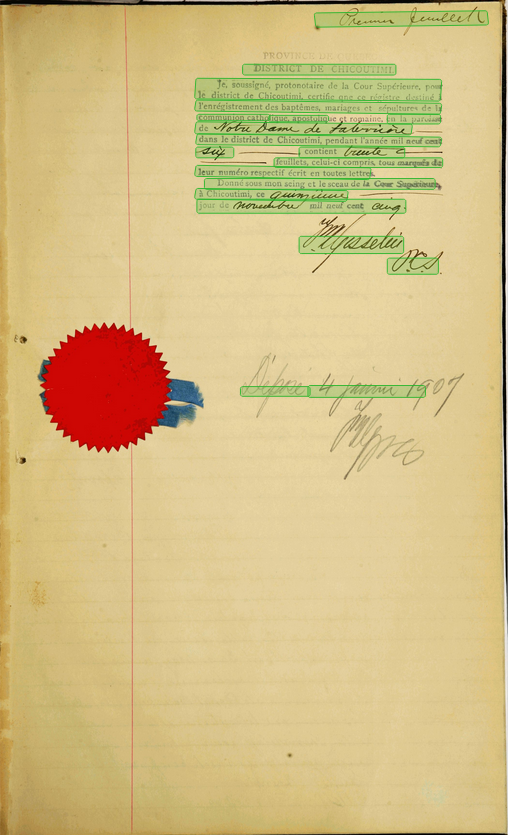}
    \hfill
    \includegraphics[width=0.182\textwidth]{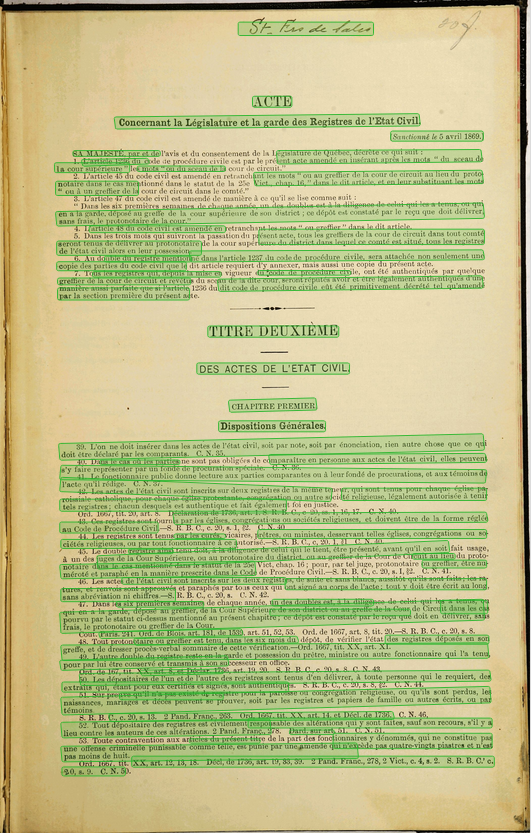}
    \caption{Illustration of different pages with various bad line ratios. From left to right: $1\%$, $5\%$, $14\%$, $30\%$, and $52\%$ of bad lines.} 
    \label{fig:bad_line_page_examples}
\end{figure*}

On Figure \ref{fig:bad_line_page_examples} there is an example image for each class. Images with ratios $1\%$ and $5\%$ show very precise detections. The page with ratio $14\%$ has some errors and since there are not many lines on that image, the bad line ratio goes up. Images with ratios $30\%$ and $52\%$ contain typed text that the line detection model has not seen during training. Understandably, the detection does not work as well on those pages, but it can easily be improved by adding them to the training set.

\begingroup
\setlength{\tabcolsep}{8pt}
\begin{table}[t]
    \centering
    \caption{Distribution of bad line classes on the images extracted from the registers of the Saguenay region}
    \label{tab:bad_line_classes_distribution}
    \begin{tabular}{lr}
    \toprule
    \textbf{Bad lines} &  \textbf{Count} \\
    \midrule
    $\leq$ 1\%  &   27,843 \\
    1$-$5\%    &   14,818 \\
    5$-$25\%   &    3,971 \\
    25$-$50\%  &     584 \\
    $>$ 50\%    &      42 \\
    \midrule
    \textbf{Total}      &   47,258 \\
    \bottomrule
    \end{tabular}
\end{table}
\endgroup


\subsection{Handwritten text recognition}
\label{sec:handwritten-text-recognition}

\subsubsection{Handwritten text recognition models}

The handwritten text recognition (HTR) is performed on the text lines detected by the Doc-UFCN model, and outputs the corresponding text. We have compared two open-source HTR engines for this task:
\begin{itemize}
    \item The Kaldi toolkit\footnote{\url{https://kaldi-asr.org/}}, which is based on a Deep Neural Network - Hidden Markov Model (DNN-HMM) model. 
    \item PyLaia\footnote{\url{https://github.com/jpuigcerver/PyLaia}}, which is a DNN model introduced in \cite{PyLaia}, that we combine with an external N-gram language model to improve performance.
\end{itemize}  

\textbf{Kaldi} model is similar to the one proposed in \cite{kaldi}. It is composed of two standard components: first, an optical model, which is trained on text line images, learns the different shapes of sub-words defined using the BPE approach \cite{sennrich2016}. During the decoding phase, it predicts sub-words probabilities in the text line using a sliding window segmentation. Second, a language model, which is trained on the text of the target documents, models linguistic information based on the frequencies of sequences of sub-words.

In our experiments, we used an optical model composed of 6 layers of convolutions with ReLU (Rectified Linear Unit) and batch normalization; 4 layers of TDNNs with ReLU and batch normalization; and an output layer with softmax. The language model is a 3-gram with Witten-Bell smoothing trained on the transcribed data only. A sub-word level language model can deal with short text lines in which the words are often split, such as we have in the BALSAC parish records, and can also predict out-of-vocabulary words, such as names, as sequences of sub-words. 

\textbf{PyLaia} model is similar to the “expert” model presented in \cite{HTR-opensource}. First, images are resized to a fixed height of 128 pixels. Data augmentation is applied on the fly and consists of random affine transformations. The model is composed of 4 convolutional layers and 3 LSTM layers. We use Leaky ReLU activation function, batch normalization and max pooling for each convolution layers. The final layer is a linear layer with softmax activation function that provides probabilities associated with each character of the vocabulary. 
The model is trained using the CTC loss function and a learning rate scheduler to ease convergence. The training is stopped after 50 epochs with no improvement.

PyLaia can be easily combined with a language model to improve its performance. We build a 6-gram character-based language model with Witten-Bell smoothing trained on training data only. It is created using the SRILM\footnote{\url{http://www.speech.sri.com/projects/srilm/}} toolkit. 

\subsubsection{Handwritten text recognition results}

Handwritten text recognition is evaluated using the Character Error Rate (CER) and Word Error Rate (WER) metrics. The CER is the edit distance (Levenshtein distance) between two strings, normalized by length. One string can be transformed into another using three edit operations: insertion, deletion and substitution of one character. The minimal number of operations needed to transform a string into another is called the edit distance. The WER works on the same principle, but at word level instead of character level.

The results are presented in Table \ref{tab:htr_results_new}. It can be observed that PyLaia is less accurate than Kaldi. However, when combined with a 6-gram character-based language model, it achieves comparable performance. 
On the test set, Kaldi achieves lower Word Error Rate, while PyLaia combined with a language model achieves lower Character Error Rate. In the end, we decided to use Kaldi, as it was already integrated in Arkindex. 

The results produced by the HTR system are critical as they greatly impact named entity recognition results. 

\begingroup
\setlength{\tabcolsep}{8pt}
\begin{table*}[t]
\centering
\caption{Handwritten text recognition evaluation for Kaldi and PyLaia. We present the character error rate (CER) and word error rate (WER) on different sets}
\label{tab:htr_results_new}
\begin{tabular}{lccccccc} \toprule
 & \multicolumn{2}{c}{\textbf{train}} & \multicolumn{2}{c}{\textbf{val}} & \multicolumn{2}{c}{\textbf{test}}\\ \cmidrule(lr){2-3}\cmidrule(lr){4-5}\cmidrule(lr){6-7}
& CER & WER & CER & WER & CER & WER \\ \midrule
\textbf{Kaldi} & 4.13 & 12.36 & 6.22 & \textbf{17.10} & 6.41 & \textbf{17.41} \\
\textbf{PyLaia (no LM)} & 2.52 & 9.67 & 6.53 & 19.75 & 6.92 & 20.41 \\
\textbf{PyLaia (with LM)} & \textbf{2.24} & \textbf{8.27} & \textbf{5.90} &  17.16 & \textbf{6.29} & 17.92 \\
\bottomrule
\end{tabular}
\end{table*}
\endgroup

\subsection{Named-entity recognition}

For the name-entity recognition step, we compared the performance of three open-source  NER libraries: Stanza, Flair and spaCy. 

\textbf{Stanza (v1.1.1)}\footnote{\url{https://stanfordnlp.github.io/stanza}} \cite{qi2020stanza} 
was created by the Stanford NLP group. It provides different NLP tools that can be used in a pipeline (e.g., tokenization, NER) and pre-trained neural models supporting 66 different languages, including English and French for named-entity extraction. 

\textbf{Flair (v0.7)}\footnote{\url{https://github.com/flairNLP/flair}} \cite{akbik2018coling} provides, besides state-of-the-art NLP models, contextual word and character embeddings, a feature that is now also supported by Stanza. Moreover, Flair allows stacking embeddings, on top of each other, a configuration  that we adopted in this paper.

\textbf{spaCy (v2.3.5)}\footnote{\url{https://spacy.io}} \cite{spacy} is 
explicitly designed to ease deploying models into production. The library comes with out-of-the-box support of multiple languages, including French and English. 

For each library, the pre-trained French model was fine-tuned using the annotated BALSAC data for named-entity recognition using the default values for the hyperparameters since they are usually the optimal values.

The evaluation of the named-entity extraction model on the BALSAC test set using both the manual transcription and the automatic transcription is presented in Table \ref{tab:ner_results}. The metric used for the evaluation of the NER models is computed as follows: the automatic transcription is first aligned with the ground truth at character level, by minimizing the Levenshtein distance between them. Each entity in the ground truth is then matched with a corresponding entity in the aligned transcription, that is with the same entity tag, or an empty character string if not found. If the ratio of the edit distance between the two entities is less than 30\% of the ground truth length, the entity is considered as recognized. We apply this metric to the manual transcription, the alignment is always perfect (since both transcription are identical) and only a difference of entity tag is considered as an error. Using this metric for both the manual transcription and the automatic transcription allows evaluating the drop of performance of the NER model due to the text line detection and the handwriting recognition. Table \ref{tab:ner_results} shows that this drop in performance is rather limited, around 8\%.

\begin{table}[t]
\centering
\caption{Named-entity recognition results on the BALSAC test set both with manual transcription and automatic transcription with the Kaldi recognizer, for the three NER libraries}
\label{tab:ner_results}
\begin{tabular}{lrrrr}
\toprule
& \textbf{Text}& \textbf{Precision} & \textbf{Recall} & \textbf{F1-score} \\
\midrule
\textbf{Stanza} & manual & 77.9 & 85.5 & 81.5 \\
\textbf{Stanza} & auto. & 69.7 & 78.3 & 73.7 \\
\textbf{Flair} & manual & \textbf{93.7}  & \textbf{93.1} & \textbf{93.4} \\
\textbf{Flair} & auto. & \textbf{86.0} & \textbf{85.6} & \textbf{85.8} \\
\textbf{Spacy} & manual & 83.1 & 83.6 & 83.4 \\
\textbf{Spacy} & auto. & 74.5 & 76.7 & 75.6 \\ \bottomrule
\end{tabular}
\end{table}

\subsection{Act detection and classification}

Inside the parish registers, the individual acts and their types (birth, marriage, or death) are the semantically meaningful unit, in which entities make sense. Therefore, all the extracted text lines must be grouped and assigned to the corresponding act.


\subsubsection{Act detection}

For the act detection, we trained a second Doc-UFCN model as described in section \ref{sec:layout-analysis}, but instead of predicting text lines, the model predicts act beginnings, act centers, act ends and full acts. The results of this model are presented in the top part of Table \ref{tab:act_segmentation}. We notice that the mean AP value for the full act class is only of 38\%. This quite low value often reflects a significant number of mergers in the prediction. Indeed, when looking at the visual results, we noticed that a large number of consecutive full acts were merged in the predictions. To help the detection model to better split consecutive full acts, we trained a second detection model on enriched inputs that consider the previously detected text lines.

Since most of the acts start with a date, for example \textit{Le trente et un janvier,  mil neuf}, and there are usually no other dates mentioned in the act, we added this information to the act detection model. Our hypothesis is that the model can take advantage of the textual content to improve the detection of the acts, and in particular, the split of consecutive acts.

First, the detected lines should be classified as containing, or not, a date. For this, the lines containing a date are detected with a simple rule that counts the number of automatically recognized words that are numbers or months. In our experiments, three words seemed to be enough for the line to be considered as containing a date. Then, we need a way to integrate this information on the original input image before feeding it to the Doc-UFCN model. 

Several ways of using the lines containing a date were tried to improve the detection model:
\begin{itemize}
    \item Drawing only the contours of the detected lines classified as containing a date directly on the image;
    \item Drawing the polygons of the detected lines classified as containing a date on a fourth input channel (RGB image + polygons mask);
    \item Drawing the contours of all the detected lines directly on the image, with a different color for those classified as containing a date.
\end{itemize}

In the end, the third proposal gave the best results. Hence, the date information is encoded on the original input image by drawing detected text line polygons on the image, with those containing a date drawn with a different color. Then the enriched image with lines will be used as input to the neural net both for training and evaluation.

It is worth noting that the text line polygons and their transcriptions used to classify the text lines come from automatic processing. Indeed, they do not come from the annotations, but from the predictions of the models described in previous sections (\ref{sec:layout-analysis} and \ref{sec:handwritten-text-recognition} respectively). This way, the training and evaluation are more similar to the real use-case, where we cannot be too certain on the correctness of the lines and their transcriptions.

The first rows of Table \ref{tab:act_segmentation} show the results obtained using raw input images. The second part of the table presents the results obtained with the enriched input images, where the text lines are drawn using two distinct colors. For comparison, we also trained a dhSegment model on the same enriched input images, whose results are presented in the bottom part of Table \ref{tab:act_segmentation}.

\begingroup
\setlength{\tabcolsep}{6pt}
\begin{table*}
\centering
\caption{Act detection results obtained by Doc-UFCN and dhSegment on BALSAC test images. The top of the table shows the results obtained by Doc-UFCN trained on standard original images. The bottom presents the results obtained by Doc-UFCN and dhSegment when training a model on the enriched input images}
\label{tab:act_segmentation}
\begin{tabular}{lllrrrr}
\toprule
\textbf{Model} & \textbf{Input} & \textbf{Class} & \textbf{IoU} & \textbf{AP@.50} & \textbf{AP@.75} & \textbf{mAP} \\ \midrule
\multirow{3}{*}{\textbf{Doc-UFCN}} & \multirow{3}{*}{\textbf{Images}} & \textit{full} & \textbf{0.84} & 0.57 & 0.37 & 0.38 \\
& & \textit{start} & \textbf{0.58} & 0.86 & 0.85 & 0.76 \\
& & \textit{end} & \textbf{0.58} & 0.85 & 0.64 & 0.59 \\ \midrule
\multirow{3}{*}{\textbf{Doc-UFCN}} & \multirow{2}{*}{\textbf{Images}} & \textit{full} & 0.82 & \textbf{0.89} & \textbf{0.81} & \textbf{0.74} \\
& \multirow{2}{*}{\textbf{+ Lines}} & \textit{start} & \textbf{0.58} & \textbf{0.90} & \textbf{0.87} & \textbf{0.78} \\
& & \textit{end} & 0.54 & \textbf{0.96} & \textbf{0.73} & \textbf{0.63} \\ \midrule
\multirow{3}{*}{\textbf{dhSegment}} & \multirow{2}{*}{\textbf{Images}} & \textit{full} & 0.75 & 0.19 & 0.08 & 0.08 \\
& \multirow{2}{*}{\textbf{+ Lines}} & \textit{start} & 0.48 & 0.81 & 0.49 & 0.47 \\
& & \textit{end} & 0.41 & 0.80 & 0.48 & 0.48 \\
\bottomrule
\end{tabular}
\end{table*}
\endgroup

The Average Precision evaluates a model at object level: it represents the amount of correctly detected objects, whereas Intersection-over-Union shows the amount of correctly predicted pixels. Even if the IoU values are on average slightly better for the model trained on original images, when looking at the AP results, it is clear that the model trained on enriched input images is way better. Indeed, having a higher AP result means that the detected acts are more precise and not merged, which cannot be reflected by the IoU metric. dhSegment results are way lower than those of Doc-UFCN, especially on the full act class, which can reveal a lot of merged predicted acts.

\subsubsection{Act type classification}

\todo[inline]{Keyword matching uniquement en Français alors qu'on traite des registres français et anglais}
The last step is to retrieve the type of each act, either \textit{birth}, \textit{marriage} or \textit{death}.
Since this information is not included in the training data, the classification algorithm was kept simple. For each predicted act, the transcription is matched against a list of keywords by act type. The classification is then performed based on the matching ratio.

The list of keywords by type was selected manually by counting all the word occurrences in the training set. It is presented in Table \ref{tab:act_type_keywords}. Keywords that were considered as discriminative of an act type were kept. 

\begin{table}
\centering
    \caption{Keywords used for act classification for birth, marriage, and death acts}
    \label{tab:act_type_keywords}
    \begin{tabular}{lll}
    \toprule
      \textbf{Birth} &      \textbf{Marriage} &       \textbf{Death} \\
    \midrule
      \textit{baptisé} &      \textit{marriage} &      \textit{inhumé} \\
          \textit{née} &          \textit{bans} &   \textit{cimetière} \\
           \textit{né} &  \textit{consentement} &       \textit{corps} \\
      \textit{parrain} &      \textit{nuptiale} &      \textit{décédé} \\
     \textit{marraine} &       \textit{\textit{nuptial}} &  \textit{inhumation} \\
              &   \textit{empêchement} &     \textit{défunte} \\
              &    \textit{opposition} &             \\
              &         \textit{époux} &             \\
              &        \textit{épouse} &             \\
    \bottomrule
    \end{tabular}
\end{table}

As the training data does not include the act type, the act classification algorithm could not be properly evaluated. However, we performed a qualitative assessment and found the classification results very accurate. 
The results are obviously very dependent on the quality of the predicted transcriptions. Fortunately, as the selected keywords are very frequent in the training set, they are very often recognized without any error by the HTR system.


\paragraph{}
With this complete workflow, we have processed more than 2 million pages. Results were exported into XML files, one for each register. 
They were sent to the BALSAC team for content standardization and verification.

\section{Processing the transcriptions at BALSAC}
\label{sec:validation}
\subsection{Challenges}
A total of 44,742 files were received at BALSAC for integration in the database. Each file corresponds to a register. As mentioned before, linkage is the central step of this procedure as the construction of the database is based on linkage of civil records. Linkage is attempted on each new record using nominative information to connect it to an individual ID within the database. For example, a birth record contains information on three people: the subject of the event, the child, and his/her parents, father and mother. Each of these individuals is linked to an ID in the database, and the birth, is entered as an event, linking those three individuals. If no ID for an individual can be found in the database, it indicates that no event involving him or her was previously entered in BALSAC and a new ID is created.

It is important to note that the BALSAC database already contains Quebec Catholic marriages up to 1965. Therefore, for the linkage to work, the name of each member of the couple represents the key information to obtain in order to connect an act to the right family. Every type of record mentions a couple, whether as parents in a birth or in the death of a single individual (child or adult) or as a couple at a marriage or at the death of one of the two spouses. In addition, other elements such as the name of the parish or the date of the event can provide helpful information to choose between homonym couples. 

For BALSAC, all these elements, up to now, were detected by human eyes. A workflow consisting of research assistants reading the act and extracting the material selected for entry (and linkage), meant that some of the validation and the standardization operations were performed while manually entering the information. In contrast, working with AI underpins the necessity to create a validation and standardization process that covers all operations since none of these can be done at the time of data reading and extraction. These are needed to ensure that the right information is entered into the database and can be used to initiate the linkage procedure.

\subsection{Act extraction}
The files sent to BALSAC are .xml documents containing the text detected by the HTR model as well as additional information regarding the structure of the text. The very first step of the BALSAC integration workflow relies on the information on the size of the text entity to remove the acts with irregular layout which can thus unambiguously be rejected. At the same time, another process combines sections of acts that are found on two separate pages based on a flag called \texttt{Act\_Class} which specifies if a text entity is tagged as an \textit{act start}, \textit{act center}, \textit{act end} or \textit{act full}, as described in Section \ref{sec:annotated-ground-truth}.

Among the files received, the decision was made to first work with the registers from Catholic parishes or missions. Some registers come from other places such as hospitals or orphanages; they are less standard in their content or presentation and will therefore be more complex to deal with. Registers in English, whether Catholic or of other denominations, were also left out for the time being. Thus, for this first round of work, 29,284 registers (66\% of the total) corresponding to 1,480,471 images (74\% of the total) were processed. Altogether 3,125,951 acts were detected by the workflow in these registers. 


\subsection{Named entity extraction and consistency checks}
One of the essential indicators on the .xml documents is the flag \texttt{Act\_Type}. Detection of this indicator starts a new step in the workflow by classifying each record among the three types of events that can possibly be recorded in the registers, i.e. birth, marriage and death. In the case of acts composed of the combination of two text entities, the type is taken from the first part or from the second part if the first one was undefined Table \ref{tab:validation_acts} gives the distribution of the acts according to type. 

\begin{table}[t]
    \centering
    \caption{Distribution of acts according to type that were detected in the 29,284 registers processed at BALSAC}
    \label{tab:validation_acts}
    \begin{tabular}{lrr}
    \toprule
        \textbf{Act type} & \textbf{Count} & \textbf{Percentage} \\
        \midrule
        \textbf{Birth} & 1,722,585 &  54.4\% \\
        \textbf{Death} & 769,804 & 24.3\% \\
        \textbf{Marriage} & 382,234 & 12.1\% \\
        \textbf{Undefined} & 291,120 & 9.2\% \\
        \midrule
        \textbf{Total} & 3,165,743 & 100.0\% \\
        \bottomrule
    \end{tabular}
\end{table}

As mentioned above, all Catholic marriages up to 1965 are already entered and linked in the database. Therefore, linkage efforts are concentrated on births and deaths which represent 79\% of all records in our dataset. The undefined category contains acts with a shape and size that could correspond to a valid act but that could not be classified in a type. This can be explained by a bad condition of the register (pages ripped or stained that prevent reading certain words) or by a bad transcription of the text preventing the detection of the keywords necessary for classification.

Each \texttt{Act\_Type} is associated to a mould, based on the structure of a typical act for that event, which helps to define the information given in each piece of text relayed by the .xml. For example, in a birth record, names always follow the same order, thus, once a text marked by \texttt{Act\_Type}: Birth is divided into its mould, it is known that the first name entity is that of the child followed by that of the father and the mother. As death records present a very different structure depending on whether they relate to the death of a single or a married individual, two separate moulds were defined. To distinguish the two types of death, keywords are used. In the death certificates of single individuals, words like \textit{fils de}, \textit{fille de}, \textit{enfant de} are found. The declared age at death can also be a good indicator when it is low (below possible age at marriage) or contains words such as weeks and hours or even expressions such as a few instants or a few minutes. For married individuals, the death record would contain words such as \textit{époux}, \textit{épouse}, \textit{mari}, \textit{femme}, \textit{veuf}, \textit{veuve}.

The division of the text into the mould triggers a sequence of procedures to validate each field and standardize the information it contains. First, the validation process verifies that all fields are filled in. Some fields are set as optional whereas the others are mandatory, as presented in Table \ref{tab:fields_records}. For all moulds, the mandatory fields are the names of the main couple in the act, since, as stated above, they represent the basic information required to successfully complete the linkage. If an act contains an empty mandatory field, it is discarded as erroneous and will not go further into the integration procedure. For now, those acts are flagged as \textit{invalid} and have not been reprocessed, but they represent interesting cases to improve and strengthen the entire process. 

\begin{table*}[t]
\centering
\caption{Mandatory \cmark and optional fields (\cmark) for each type of records}
\begin{tabular}{llccc}
\toprule
& \textbf{Field} & \textbf{Birth} &\textbf{\begin{tabular}[c]{@{}c@{}}Death\\ (married person)\end{tabular}} & \textbf{\begin{tabular}[c]{@{}c@{}}Death\\ (single person)\end{tabular}} \\
\midrule
\multirow{2}{*}{\textbf{Record}} & Date of the record & \cmark & \cmark & \cmark \\
 & Place of the record & (\cmark) & (\cmark) & (\cmark) \\
\midrule
\multirow{3}{*}{\textbf{Subject}} & Name of the subject & \cmark & \cmark & \cmark \\
 & Date of the event & (\cmark) & (\cmark) & (\cmark) \\
 & Age of the subject & (\cmark) & (\cmark) & (\cmark) \\
 \midrule
\multirow{3}{*}{\textbf{Father}} & Name of the father & \cmark &  & \cmark \\
 & Occupation of the father & (\cmark) &  & (\cmark) \\
 & Residency of the father & (\cmark) &  & (\cmark) \\
\midrule
\multirow{3}{*}{\textbf{Mother}} & Name of the mother & \cmark &  & \cmark \\
 & Occupation of the mother & (\cmark) &  & (\cmark) \\
 & Residence of the mother & (\cmark) &  & (\cmark) \\
 \midrule
\multirow{3}{*}{\textbf{Spouse}} & Name of the spouse &  & \cmark &  \\
 & Occupation of the spouse &  & (\cmark) &  \\
 & Residency of the spouse &  & (\cmark) &  \\
 \midrule
\multirow{3}{*}{\textbf{Godfather}} & Name of the godfather & (\cmark) &  &  \\
 & Occupation of the godfather & (\cmark) &  &  \\
 & Residence of the godfather & (\cmark) &  &  \\
 \midrule
\multirow{3}{*}{\textbf{Godmother}} & Name of the godmother & (\cmark) &  &  \\
 & Occupation of the godmother & (\cmark) &  &  \\
 & Residence of the godmother & (\cmark) &  & \\
 \midrule
\textbf{Witnesses} & Names of the witnesses &  & (\cmark) & (\cmark) \\
 \bottomrule
\end{tabular}
\label{tab:fields_records}
\end{table*}

\subsection{Standardization}
\begin{figure}[t]
    \centering
    \includegraphics[width=\linewidth]{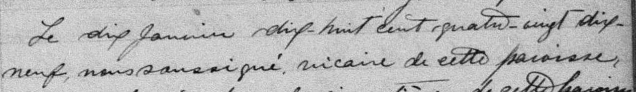}
    \includegraphics[width=\linewidth]{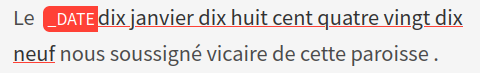}
    \caption{Example of presentation of a date in an act.}
    \label{fig:date_in_act}
\end{figure}

The next step is the standardization of the information. Since the HTR system conveys the information literally, there is a need to translate the information into the format employed in the database. Date is a perfect field to exemplify the necessity of standardization: the .xml outputs the information in words, as it appears in the original source, but it needs to be entered into the database as a date field (numeric field). Thus, an algorithm changes each word detected into a number. This algorithm needs to be flexible and comprehensive as French language has a complex linguistic structure to write numbers (see Figure \ref{fig:date_in_act}). For instance, the year 1899 written in full would be dix huit cent quatre-vingt-dix-neuf, which would literally be numerically translated into: 10-8-100-4-20-10-9. Therefore, this system of numeric-translation is combined with multiple verification criteria to make sure that the result is consistent with the other information found in the act or in the register and can thus be processed in the linkage procedure. Lastly, some dates are expressed relative to another one which is precise. For instance, the date of an event is often mentioned in relation to the date of the record which is specified at the beginning of the act. In these cases, the precise date of registration is used to calculate a precise date for the event.

Name entity is the other mandatory field that needs to be standardized. As the NER model extracts a full name, it is necessary to separate first and last names in order to perform linkage. The algorithm is case sensitive and detects spaces and hyphens. It uses this information to separate or join (when on two lines) the words included in the name field. Once the words have been separated, each one is compared to an existing thesaurus comprising more than 1,8 million names. Each of these names appears at least once in the database and, based on their appearances as first or last name, a probability of being one or the other is calculated for each one of them. This tool is used to classify each word as a first or last name. However, it should be noted that the first word in a sequence of names is always treated as a first name. In addition, the \texttt{Act\_Type} can provide some cues on the way name will be written in the source. For example, the subject of a birth act will normally be named without its last name, but with all its given first names. On the other hand, in a death act, only the usual first name will be written, but the last name will be specified. 

The variability in how names may appear in the source necessitates having all-encompassing systems capable of identifying different versions of the same information and reformatting them all into the same standardized format. BALSAC’s thesaurus is a very complete guide that has been consolidated over the last 50 years. It contains phonetic and writing variations for the same name, such as \textit{Édouard} as an equivalent of \textit{Edward}, or \textit{William} as a homonym of \textit{Guillaume}. Nonetheless, with the reading errors induced by the HTR system, it was also necessary to develop a permutation algorithm that focuses on the similarity of the handwritten characters. 

\begin{figure*}[t]
    \centering
    \includegraphics[width=\textwidth]{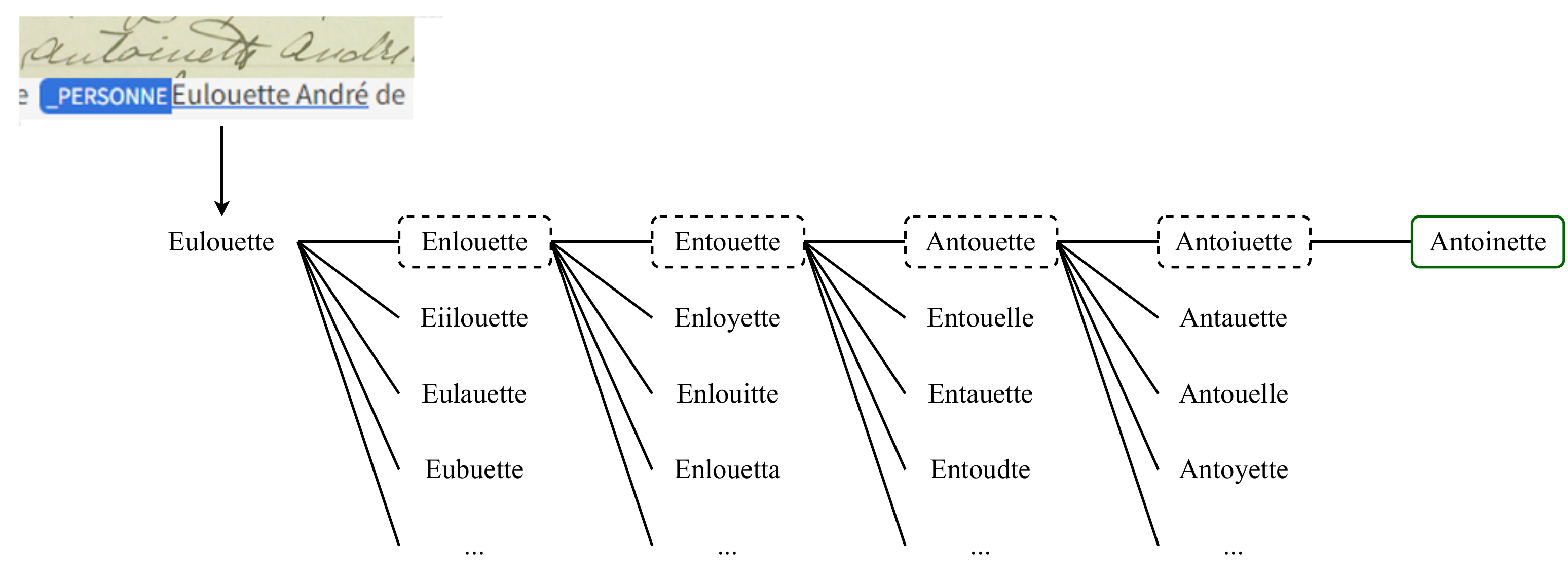}
    \caption{Illustration of how names are corrected based on a distance to BALSAC's thesaurus.}
    \label{fig:name_postprocessing}
\end{figure*}

As shown on Figure \ref{fig:name_postprocessing}, each name judged to be erroneous is compared to the thesaurus and a list of all possible names in a reasonable range of the original using the Levenshtein distance is kept as reference. Using a custom algorithm based on A*, permutations are made and scored by the visual similarity of the characters. The best scored name (or names in some instances) from the reference pool is then chosen as the most probable.

\begin{table*}[t]
\centering
\caption{Distribution of birth and death records according to their validity status}
\label{tab:validity_status}
\begin{tabular}{lrrrr}
    \toprule
    \multicolumn{1}{c}{} & \multicolumn{2}{c}{\textbf{Birth}} & \multicolumn{2}{c}{\textbf{Death}} \\
    & \multicolumn{1}{c}{Count} & \multicolumn{1}{c}{Percentage} & \multicolumn{1}{c}{Count} & \multicolumn{1}{c}{Percentage} \\
    \midrule
    \textbf{Valid} & 1,301,204 & 75.5\% & 531,405& 69.0\% \\
    \textbf{Fusion} & 291,874 & 16.9\% & 140,140 & 18.2\% \\
    \textbf{Invalid} & 122,052 & 7.1\% & 68,851 & 8.9\% \\
\textbf{Special cases} & 7,455 & 0.4\% & 24,408 & 3.2\% \\
    \midrule
    \textbf{Total} & 1,722,585 & 100.0 & 769,804 & 100.0\\
\bottomrule
\end{tabular}
\end{table*}

In parallel, another step of validation is carried out to confirm that the processed text corresponds to a single act, by verifying if no information is duplicated. For instance, it detects if a text entity contains more than one date written in full, which would suggest a fusion of two acts. A fusion happens when two acts were not segmented properly and were identified in the .xml as the same text entity. Since this situation will complicate any other validation process, those text entities are excluded from the linkage procedure and flagged as \textit{fusion}.

Then, a final step of validation is run to exclude all acts that correspond to special cases. Those acts have a valid structure but their content is too unique to be run in the current logic of the linkage procedure. Those special cases are detected with a list of keywords. Those keywords are grouped into three large categories: words identifying members of an indigenous community of Canada; words encompassing non-identified subjects; words related to immigration. The first group points to acts including a member of an indigenous community. Since in those acts, at the time, the custom was to write only the first name, accompanied by a word identifying the person as indigenous, the name is too partial to complete the linkage procedure. The second group concerns acts in which one or two parents are unknown, such as in the birth record of a foundling or an illegitimate child. Since the couple is missing or incomplete, linkage is impossible. The third group mostly focuses on death acts recorded in immigrant boats. Since the family is most likely not existing in the database, linkage would not work either. Thus, for now, those three categories of acts are flagged as \textit{special cases} and are put aside until we create a specific linkage algorithm taking into account the specific features of their content.

At the end of the process, all acts (or more precisely text entities) that do not comply with the validation criteria are not submitted to linkage and are classified as \textit{invalid} or \textit{fusion} or \textit{special cases}. Those acts are problematic either because they were not detected properly, had segmentation issues, or because the text in itself is not an act or is too unique to be processed automatically. Some of them could be processed again later as work is ongoing to improve aspects of the process such act segmentation. Others will necessitate manual intervention in order to correct or complete the transcription before they can be processed again. Table \ref{tab:validity_status} shows the distribution of the acts according to their validity status. It indicates that 74\% of records are deemed valid meaning that linkage operations can be undertaken to integrate them in the BALSAC database. For now, the linkage has a fully automatic phase and a computer-assisted manual phase. One of the objectives of the next few years at BALSAC will be to develop an AI-based algorithm for linking the civil records.

\section{Conclusion}

\label{sec:conclusion}


In this article, we have described a complete workflow designed for automatic information extraction in Quebec parish records at large scale. 
A succession of machine learning models were used for page classification, textline detection, handwritten text recognition, named entity recognition, and act detection and classification. 
We designed an unsupervised metric for quality estimation to reject incomplete or invalid extracted records before adding them in the BALSAC database.  

Working on this project has brought to light three main challenges and opened new questions. 

While setting up the complete workflow, we found out that the main steps (text line detection, handwritten text recognition, named entity recognition) were not the most difficult steps: these tasks are well studied in our field, and many off-the-shelf models are available. However, the difficulties lie in more specific tasks, such as act detection. 

During the validation procedure, we learned that unsupervised quality estimation is a challenge. We believe that it should be addressed more frequently in the field of automatic document recognition. 
This also raises the issue of how to benefit from this metric while training the different machine learning models.

On BALSAC side, it also raises challenges for the integration of data from multiple versions of transcription results while preserving the integrity of the database as well as previously added data and linkage results.

Working with AI opens up exciting new possibilities for the development of historical databases and for their users, but it also brings new kinds of limitations or biases in the data that translate into new challenges for users. As mentioned above, many steps in the BALSAC workflow result in the removal, at least temporarily, of part of the data. This means that users work with datasets that are incomplete for different reasons which in many cases have more to do with the form than the content of the data and which are quite complex to pin down and define. In the case of i-BALSAC, this situation applies in the context of the transcription of parish records where the rejected data is randomly distributed over the range of years and regions it covers. These limitations certainly do not outweigh the benefits but they should be taken into account when considering the contribution of this technology to database development. 


\section*{Acknowledgments}
The i-BALSAC project is supported by the Canadian Foundation for Innovation through its Cyberinfrastructure Initiative. 
Mélodie Boillet is partly funded by the CIFRE ANRT grant No. 2020/0390.

\bibliographystyle{splncs04}
\bibliography{bibliography}

\begin{thebibliography}{10}
\providecommand{\url}[1]{\texttt{#1}}
\providecommand{\urlprefix}{URL }
\providecommand{\doi}[1]{https://doi.org/#1}

\bibitem{abadie2022-ner-survey}
Abadie, N., Carlinet, E., Chazalon, J., Dum{\'e}nieu, B.: {A Benchmark of Named
  Entity Recognition Approaches in Historical Documents Application to 19th
  Century French Directories}. In: {Document Analysis Systems}. pp. 445--460
  (2022)

\bibitem{akbik2018coling}
Akbik, A., Blythe, D., Vollgraf, R.: {Contextual String Embeddings for Sequence
  Labeling}. In: {Proceedings of the 27th International Conference on
  Computational Linguistics}. pp. 1638--1649 (Aug 2018)

\bibitem{dhsegment}
Ares~Oliveira, S., Seguin, B., Kaplan, F.: {dhSegment: A Generic Deep-learning
  Approach for Document Segmentation}. In: {16th International Conference on
  Frontiers in Handwriting Recognition (ICFHR)}. pp. 7--12 (Aug 2018)

\bibitem{kaldi}
Arora, A., Chang, C.C., Rekabdar, B., BabaAli, B., Povey, D., Etter, D., Raj,
  D., Hadian, H., Trmal, J., Garcia, P., et~al.: {Using ASR Methods for OCR}.
  In: {15th International Conference on Document Analysis and Recognition}. pp.
  663--668 (Sep 2019)

\bibitem{bluche2016}
Bluche, T., Louradour, J., Messina, R.O.: {Scan, Attend and Read: End-to-End
  Handwritten Paragraph Recognition with {MDLSTM} Attention}. In:
  {International Conference on Document Analysis and Recognition}. pp.
  1050--1055 (Nov 2017). \doi{10.1109/ICDAR.2017.174}

\bibitem{boillet2021}
Boillet, M., Maarand, M., Paquet, T., Kermorvant, C.: {Including Keyword
  Position in Image-Based Models for Act Segmentation of Historical Registers}.
  In: {6th International Workshop on Historical Document Imaging and
  Processing}. p. 31–36 (Sep 2021). \doi{10.1145/3476887.3476905}

\bibitem{boillet2020}
Boillet, M., Kermorvant, C., Paquet, T.: {Multiple Document Datasets
  Pre-training Improves Text Line Detection With Deep Neural Networks}. In:
  {25th International Conference on Pattern Recognition}. pp. 2134--2141 (Jan
  2020)

\bibitem{breunig2000lof}
Breunig, M.M., Kriegel, H.P., Ng, R.T., Sander, J.: {LOF: Identifying
  Density-based Local Outliers}. In: {2000 ACM SIGMOD International Conference
  on Management of Data}. pp. 93--104 (2000)

\bibitem{capobianco2019}
Capobianco, S., Marinai, S.: {Deep Neural Networks for Record Counting in
  Historical Handwritten Documents}. {Pattern Recognition Letters}
  \textbf{119},  103--111 (2017). \doi{10.1016/j.patrec.2017.10.023}

\bibitem{Carbonell2020}
Carbonell, M., Fornés, A., Villegas, M., Lladós, J.: {A Neural Model for Text
  Localization, Transcription and Named Entity Recognition in Full Pages}.
  {Pattern Recognition Letters}  \textbf{136},  219--227 (May 2020).
  \doi{10.1016/j.patrec.2020.05.001}

\bibitem{carbonell2018}
Carbonell, M., Villegas, M., Forn{\'{e}}s, A., Llad{\'{o}}s, J.: Joint
  recognition of handwritten text and named entities with a neural end-to-end
  model. In: 2018 13th IAPR International Workshop on Document Analysis Systems
  (DAS). pp. 399--404. IEEE Computer Society, Los Alamitos, CA, USA (apr 2018).
  \doi{10.1109/DAS.2018.52},
  \url{https://doi.ieeecomputersociety.org/10.1109/DAS.2018.52}

\bibitem{constum2022}
Constum, T., Kempf, N., Paquet, T., Tranouez, P., Chatelain, C., Br{\'e}e, S.,
  Merveille, F.: {Recognition and Information Extraction in Historical
  Handwritten Tables: Toward Understanding Early 20th Century Paris Census}.
  In: {Document Analysis Systems}. pp. 143--157 (2022)

\bibitem{dan}
Coquenet, D., Chatelain, C., Paquet, T.: {DAN: a Segmentation-free Document
  Attention Network for Handwritten Document Recognition} (2022).
  \doi{10.48550/ARXIV.2203.12273}

\bibitem{imagenet}
Deng, J., Dong, W., Socher, R., Li, L.J., Li, K., Li, F.F.: {ImageNet: a
  Large-Scale Hierarchical Image Database}. In: {2009 IEEE Conference on
  Computer Vision and Pattern Recognition}. pp. 248--255 (Jun 2009).
  \doi{10.1109/CVPR.2009.5206848}

\bibitem{douzon2022}
Douzon, T., Duffner, S., Garcia, C., Espinas, J.: {Improving Information
  Extraction on Business Documents with Specific Pre-training Tasks}. In:
  {Document Analysis Systems}. pp. 111--125 (2022)

\bibitem{embley2018}
Embley, D.W., Nagy, G.: {Green Interaction for Extracting Family Information
  from OCR'd Books}. In: {2018 13th IAPR International Workshop on Document
  Analysis Systems}. pp. 127--132 (2018). \doi{10.1109/DAS.2018.58}

\bibitem{IEHHR2017}
Fornés, A., Romero, V., Baró, A., Toledo, J.I., Sánchez, J.A., Vidal, E.,
  Lladós, J.: {ICDAR2017 Competition on Information Extraction in Historical
  Handwritten Records}. In: 2017 14th IAPR International Conference on Document
  Analysis and Recognition. vol.~01, pp. 1389--1394 (2017).
  \doi{10.1109/ICDAR.2017.227}

\bibitem{readbad2017}
Grüning, T., Labahn, R., Diem, M., Kleber, F., Fiel, S.: {READ-BAD: A New
  Dataset and Evaluation Scheme for Baseline Detection in Archival Documents}.
  In: {13th International Workshop on Document Analysis Systems}. pp. 351--356
  (May 2017)

\bibitem{gruning2018}
Grüning, T., Leifert, G., Strauß, T., Labahn, R.: {A Two-Stage Method for
  Text Line Detection in Historical Documents}. In: {International Journal on
  Document Analysis and Recognition}. vol.~22, pp. 285--302 (Sep 2019).
  \doi{10.1007/s10032-019-00332-1}

\bibitem{guo2017}
Guo, C., Pleiss, G., Sun, Y., Weinberger, K.Q.: {On Calibration of Modern
  Neural Networks}. In: {International Conference on Machine Learning} (2017)

\bibitem{resnet}
He, K., Zhang, X., Ren, S., Sun, J.: {Deep Residual Learning for Image
  Recognition}. In: {2016 IEEE Conference on Computer Vision and Pattern
  Recognition}. pp. 770--778 (Jun 2016). \doi{10.1109/CVPR.2016.90}

\bibitem{spacy}
Honnibal, M., Montani, I., Van~Landeghem, S., Boyd, A.: {spaCy:
  Industrial-strength Natural Language Processing in Python} (2020).
  \doi{10.5281/zenodo.1212303}, \url{https://doi.org/10.5281/zenodo.1212303}

\bibitem{transkribus}
Kahle, P., Colutto, S., Hackl, G., Mühlberger, G.: {Transkribus - A Service
  Platform for Transcription, Recognition and Retrieval of Historical
  Documents}. In: {2017 14th IAPR International Conference on Document Analysis
  and Recognition}. vol.~04, pp. 19--24 (Nov 2017).
  \doi{10.1109/ICDAR.2017.307}

\bibitem{das-pageclasification}
Kiss, M., Koh{\'{u}}t, J., Benes, K., Hradis, M.: {Importance of Textlines in
  Historical Document Classification}. In: {Document Analysis Systems}. pp.
  158--170 (2022)

\bibitem{lang2018}
{Lang}, E., {Puigcerver}, J., {Toselli}, A.H., {Vidal}, E.: {Probabilistic
  Indexing and Search for Information Extraction on Handwritten German Parish
  Records}. In: {2018 16th International Conference on Frontiers in Handwriting
  Recognition}. pp. 44--49 (2018). \doi{10.1109/ICFHR-2018.2018.00017}

\bibitem{liu2008isolation}
Liu, F.T., Ting, K.M., Zhou, Z.H.: {Isolation Forest}. In: {2008 Eighth IEEE
  International Conference on Data Mining}. pp. 413--422 (2008)

\bibitem{gcnn_ie}
Liu, X., Gao, F., Zhang, Q., Zhao, H.: {Graph Convolution for Multimodal
  Information Extraction from Visually Rich Documents}. In: {Proceedings of the
  2019 Conference of the North American Chapter of the Association for
  Computational Linguistics: Human Language Technologies, Volume 2 (Industry
  Papers)}. pp. 32--39 (Jun 2019). \doi{10.18653/v1/N19-2005}

\bibitem{HTR-opensource}
Maarand, M., Beyer, Y., K{\aa}sen, A., Fosseide, K.T., Kermorvant, C.: A
  comprehensive comparison of open-source libraries for handwritten text
  recognition in norwegian. In: {Document Analysis Systems}. pp. 399--413
  (2022)

\bibitem{camembert}
Martin, L., Muller, B., Ortiz~Su{\'a}rez, P.J., Dupont, Y., Romary, L., de~la
  Clergerie, {\'E}., Seddah, D., Sagot, B.: {CamemBERT: a Tasty {F}rench
  Language Model}. In: {58th Annual Meeting of the Association for
  Computational Linguistics}. pp. 7203--7219 (Jul 2020)

\bibitem{monnier2020docExtractor}
Monnier, T., Aubry, M.: {docExtractor: An off-the-shelf historical document
  element extraction}. In: {International Conference on Frontiers in
  Handwriting Recognition} (2020)

\bibitem{monroc2022-ner-survey}
Monroc, C.B., Miret, B., Bonhomme, M.L., Kermorvant, C.: {A Comprehensive Study
  of Open-Source Libraries for Named Entity Recognition on Handwritten
  Historical Documents}. In: {Document Analysis Systems}. pp. 429--444 (2022)

\bibitem{nion2013}
Nion, T., Menasri, F., Louradour, J., Sibade, C., Retornaz, T., Métaireau,
  P.Y., Kermorvant, C.: {Handwritten Information Extraction from Historical
  Census Documents}. In: {2013 12th International Conference on Document
  Analysis and Recognition}. pp. 822--826 (2013). \doi{10.1109/ICDAR.2013.168}

\bibitem{scikit-learn}
Pedregosa, F., Varoquaux, G., Gramfort, A., Michel, V., Thirion, B., Grisel,
  O., Blondel, M., Prettenhofer, P., Weiss, R., Dubourg, V., Vanderplas, J.,
  Passos, A., Cournapeau, D., Brucher, M., Perrot, M., Duchesnay, E.:
  {Scikit-learn: Machine Learning in Python}. {Journal of Machine Learning
  Research}  \textbf{12},  2825--2830 (2011)

\bibitem{prieto2020}
Prieto, J.R., Bosch, V., Vidal, E., Stutzmann, D., Hamel, S.: {Text Content
  Based Layout Analysis}. In: {2020 17th International Conference on Frontiers
  in Handwriting Recognition}. pp. 258--263 (Sep 2020).
  \doi{10.1109/ICFHR2020.2020.00055}

\bibitem{PyLaia}
Puigcerver, J.: {Are Multidimensional Recurrent Layers Really Necessary for
  Handwritten Text Recognition?} In: {2017 14th IAPR International Conference
  on Document Analysis and Recognition}. vol.~01, pp. 67--72 (2017).
  \doi{10.1109/ICDAR.2017.20}

\bibitem{qi2020stanza}
Qi, P., Zhang, Y., Zhang, Y., Bolton, J., Manning, C.D.: {Stanza: A Python
  Natural Language Processing Toolkit for Many Human Languages}. In: {58th
  Annual Meeting of the Association for Computational Linguistics: System
  Demonstrations}. pp. 101--108 (Jul 2020). \doi{10.18653/v1/2020.acl-demos.14}

\bibitem{InstaDeep-JointHTRNER}
Rouhou, A.C., Dhiaf, M., Kessentini, Y., Salem, S.B.: {Transformer-based
  Approach for Joint Handwriting and Named Entity Recognition in Historical
  Document}. {Pattern Recognition Letters}  \textbf{155},  128--134 (2022).
  \doi{10.1016/j.patrec.2021.11.010}

\bibitem{sennrich2016}
Sennrich, R., Haddow, B., Birch, A.: {Neural Machine Translation of Rare Words
  with Subword Units}. In: {Annual Meeting of the Association for Computational
  Linguistics} (2016)

\bibitem{das-pageclassification-comp-2022}
Seuret, M., Nicolaou, A., Rodr{\'i}guez-Salas, D., Weichselbaumer, N.,
  Stutzmann, D., Mayr, M., Maier, A., Christlein, V.: {ICDAR 2021 Competition
  on Historical Document Classification}. In: Llad{\'o}s, J., Lopresti, D.,
  Uchida, S. (eds.) {International Conference on Document Analysis and
  Recognition}. pp. 618--634 (2021)

\bibitem{diva2016}
Simistira, F., Seuret, M., Eichenberger, N., Garz, A., Liwicki, M., Ingold, R.:
  {DIVA-HisDB: A Precisely Annotated Large Dataset of Challenging Medieval
  Manuscripts}. In: {15th International Conference on Frontiers in Handwriting
  Recognition}. pp. 471--476 (Oct 2016). \doi{10.1109/ICFHR.2016.0093}

\bibitem{tarride2021}
Tarride, S., Lemaitre, A., Co{\"u}asnon, B., Tardivel, S.: {Combination of Deep
  Neural Networks and Logical Rules for Record Segmentation in Historical
  Handwritten Registers using Few Examples}. {International Journal on Document
  Analysis and Recognition}  \textbf{24},  77--96 (2021).
  \doi{10.1007/s10032-021-00362-8}

\bibitem{tarride2022}
Tarride, S., Lemaitre, A., Co{\"u}asnon, B., Tardivel, S.: {A Comparative Study
  of Information Extraction Strategies Using an Attention-Based Neural
  Network}. In: {Document Analysis Systems}. pp. 644--658 (2022)

\bibitem{walton2022}
Walton, S., Livermore, L., Bánki, O., N. Cubey, R.W., Drinkwater, R.,
  Englund, M., Goble, C., Groom, Q., Kermorvant, C., Rey, I., M Santos, C.,
  Scott, B., R. Williams, A., Wu, Z.: Landscape analysis for the specimen data
  refinery. Research Ideas and Outcomes  \textbf{6},  e57602 (2020).
  \doi{10.3897/rio.6.e57602}, \url{https://doi.org/10.3897/rio.6.e57602}

\bibitem{Wang2021}
Wang, J., Liu, C., Jin, L., Tang, G., Zhang, J., Zhang, S., Wang, Q., Wu, Y.,
  Cai, M.: {Towards Robust Visual Information Extraction in Real World: New
  Dataset and Novel Solution}. In: {Proceedings of the AAAI Conference on
  Artificial Intelligence} (2021)

\bibitem{wigington2018}
Wigington, C., Tensmeyer, C., Davis, B., Barrett, W., Price, B., Cohen, S.:
  {Start, Follow, Read: End-to-End Full-Page Handwriting Recognition}. In:
  {ECCV 2018: 15th European Conference}. p. 372–388 (2018).
  \doi{10.1007/978-3-030-01231-1\_23}

\bibitem{layoutlm}
Xu, Y., Li, M., Cui, L., Huang, S., Wei, F., Zhou, M.: {LayoutLM: Pre-training
  of Text and Layout for Document Image Understanding}. In: {26th ACM SIGKDD
  International Conference on Knowledge Discovery \& Data Mining}. p.
  1192–1200 (Aug 2020)

\bibitem{Yu2021PICKPK}
Yu, W., Lu, N., Qi, X., Gong, P., Xiao, R.: {PICK: Processing Key Information
  Extraction from Documents using Improved Graph Learning-Convolutional
  Networks}. {2020 25th International Conference on Pattern Recognition} pp.
  4363--4370 (2021)

\end{thebibliography}
\end{document}